\documentclass{article}
\usepackage{ijcai25}

\usepackage{times}
\usepackage{soul}
\usepackage{url}
\usepackage[hidelinks]{hyperref}
\usepackage[utf8]{inputenc}
\usepackage[small]{caption}
\usepackage{graphicx}
\usepackage{amsmath}
\usepackage{amsthm}
\usepackage{booktabs}
\usepackage{algorithm}
\usepackage{algorithmic}
\usepackage[switch]{lineno}

\usepackage{amssymb}
\usepackage{subfigure}
\usepackage{makecell}
\usepackage{multirow}
\usepackage{arydshln}

\newtheorem{theorem}{Theorem}

\title{Partial Label Clustering}

\author{%
  Yutong Xie\textsuperscript{1},
  \quad Fuchao Yang\textsuperscript{2},
  \quad Yuheng Jia
  \textsuperscript{3,4}\thanks{Corresponding author} \\
  \affiliations
  \textsuperscript{1} Chien-Shiung Wu College, Southeast University, Nanjing 210096, China\\
  \textsuperscript{2} College of Software Engineering, Southeast University, Nanjing 210096, China\\
   \textsuperscript{3} School of Computer Science and Engineering, Southeast University, Nanjing 210096, China\\
  \textsuperscript{4} Key Laboratory of New Generation Artificial Intelligence Technology and Its \\ Interdisciplinary Applications (Southeast University), Ministry of Education, China
  \\
  \emails
  ytxie@seu.edu.cn, yangfc@seu.edu.cn, yhjia@seu.edu.cn
}

\begin{document}

\maketitle

\begin{abstract}
    Partial label learning (PLL) is a significant weakly supervised learning framework, where each training example corresponds to a set of candidate labels and only one label is the ground-truth label. For the first time, this paper investigates the partial label clustering problem, which takes advantage of the limited available partial labels to improve the clustering performance. Specifically, we first construct a weight matrix of examples based on their relationships in the feature space and disambiguate the candidate labels to estimate the ground-truth label based on the weight matrix. Then, we construct a set of must-link and cannot-link constraints based on the disambiguation results. Moreover, we propagate the initial must-link and cannot-link constraints based on an adversarial prior promoted dual-graph learning approach. Finally, we integrate weight matrix construction, label disambiguation, and pairwise constraints propagation into a joint model to achieve mutual enhancement. We also theoretically prove that a better disambiguated label matrix can help improve clustering performance. Comprehensive experiments demonstrate our method realizes superior performance when comparing with state-of-the-art constrained clustering methods, and outperforms PLL and semi-supervised PLL methods when only limited samples are annotated. The code is publicly available at \url{https://github.com/xyt-ml/PLC}.
\end{abstract}

\section{Introduction}
Partial label learning (PLL) \cite{TIAN2023708,jia2023complementary,10667671,jiang2024fairmatch,KDD25} is a significant weakly supervised learning framework, where each training example corresponds to a set of candidate labels, but only one of which is the ground-truth label. This form of weak supervision appears in a variety of real-world scenarios, e.g., web mining \cite{2010Learning}, multimedia content analysis \cite{7968363}, and ecoinformatics \cite{Liu2012ACM}. Previous studies in PLL mainly focus on learning a multi-class classifier based on the available weak supervision \cite{9573413,NEURIPS2023_6b97236d,jia2024long}.


 However, in many real-world scenarios, obtaining enough training examples with candidate labels may be time and resource costly as it requires a large amount of manual annotation, while sufficient unlabeled samples are readily obtainable. Given the effectiveness of clustering methods in dealing with unlabeled samples, we believe that utilizing limited partial labeled samples and sufficient unlabeled samples for clustering is an effective solution for this scenario. However, how to effectively exploit the partial labels for clustering is still under-explored. The ambiguity of the candidate labels prevents existing constrained clustering methods \cite{2018Semi,Jia2020ConstrainedCW} from directly utilizing the ground-truth labels. A naive method is to adopt label disambiguation strategies \cite{9573413,NEURIPS2023_6b97236d} to estimate the ground-truth labels, and then incorporate the estimated labels to constrained clustering methods. However, because of the limited effectiveness of existing methods in disambiguation, the performance improvement observed in experiments is marginal.

To address the above issue, we propose a novel method, named \textbf{PLC} (\textbf{P}artial \textbf{L}abel \textbf{C}lustering), to take advantage of the available partial labels to construct a clustering model. Specifically, we first establish a weight matrix for examples based on their similarity in the feature space and perform label disambiguation based on the weight matrix to obtain label information. Then we use the label information to establish initial pairwise constraints for the examples, i.e., must-link and cannot-links and propagate the initial inaccurate pairwise constraints through the weight matrix to obtain dense and precise pairwise constraints. Moreover, these two types of constraints form an adversarial relationship which are augmented accordingly. The augmented pairwise constraints are also used to adjust the weight matrix, ultimately improving the quality of the weight matrix. Finally, we use the optimized weight matrix for spectral clustering (SC) to obtain clustering results. The entire model is formulated as a joint optimization problem and solved by alternating optimization. More importantly, we also theoretically prove that a better disambiguated label matrix can improve clustering performance.



Our contributions can be summarized as follows:

\begin{itemize}
\item For the first time, we propose the partial label clustering problem and leverage the information from both the feature space and the label space of labeled and unlabeled samples to address this problem.
\item  We theoretically prove that a better disambiguated label matrix can improve clustering performance, which indicates that utilizing label disambiguation to improve clustering performance is effective. 
\item Extensive experiments demonstrate that the proposed PLC method achieves superior performances than the constrained clustering, PLL and semi-supervised PLL methods.
\end{itemize}

\section{Related Work}
In this section, we briefly introduce the preliminary work and studies related to our PLC method.

\noindent \textbf{Partial Label Learning.} In PLL, we cannot access the ground-truth labels of training examples because the ambiguity of candidate labels poses a significant challenge. The main approach to solve this problem is to disambiguate the candidate label set. For example, some methods establish a parameter model based on the maximum likelihood criterion \cite{Liu2012ACM,Jin2002LearningWM} or maximum margin criterion \cite{Nguyen2008ClassificationWP}, and identify the ground-truth label through iterative optimization. \cite{Hllermeier2005LearningFA} treats candidate labels equally and makes predictions through weighted voting based on the candidate labels of the example's neighbors. \cite{inproceedings,9573413} construct the weighted graph of the feature space of examples for disambiguation, which utilizes the similarity of examples. \cite{2019Partial,NEURIPS2023_6b97236d} construct the dissimilarity matrix for candidate labels to guide label disambiguation. Although the above PLL methods have achieved satisfactory outcomes, they need a large number of labeled examples for training to achieve the expected results, and perform poorly with a fewer number of labeled examples. To solve this problem, we introduce a constrained clustering model, achieving superior performance with a limited number of labeled examples and many easily available unlabeled examples. 

\noindent \textbf{Constrained Clustering.} As an important type of semi-supervised learning method, constrained clustering utilizes some supervision information to enhance clustering performance. Pairwise constraint is a kind of weak supervision, which represents whether two training samples belonging to the same class, i.e., must-links and cannot-links. Several methods have been proposed to exploit pairwise constraints in constrained clustering, for example, \cite{7167693} performs low dimensional embedding of pairwise constraints through non-negative matrix factorization, \cite{9178787} exploits the dissimilarity between pairwise constraints to improve the performance of constrained clustering, \cite{10007868} stacks the pairwise constraint matrix and affinity matrix into a 3-D tensor, and promotes the construction of the affinity matrix through global low-rank constraints. In addition, pairwise constraint propagation (PCP) methods \cite{Lu2010ConstrainedSC,2020Pairwise} are proposed to address the sparsity problem of pairwise constraints. However, how to effectively exploit the partial labels to construct a constrained clustering model is still under-explored.

\section{The Proposed Method}
In this section, we propose the details of PLC, including weight matrix construction, label disambiguation, and pairwise constraint propagation. Afterwards, we integrate them into a unified optimization objective and solve it through alternating optimization.
\subsection{Notations} Denote $\textbf{X}=[\textbf{\textit{x}}_1, \textbf{\textit{x}}_2,..., \textbf{\textit{x}}_n]^\top\in\mathbb{R}^{n\times d}$ as the feature matrix, where $n$ and $d$ represent the number of examples and the dimension of features. $\textbf{Y}=[\textbf{\textit{y}}_1, \textbf{\textit{y}}_2,..., \textbf{\textit{y}}_n]^\top\in\{0,1\}^{n\times q}$ represents the partial label matrix, where $q$ is the number of classes and $y_{ij}=1$ means that the $j$-th label is one of the candidate labels of the sample $\textbf{\textit{x}}_i$. Specifically, we only have the candidate labels for training examples, and for test examples, the set of candidate labels consists of all labels, i.e., $y_{test}=\textbf{1}_q$ for all test examples $\textbf{\textit{x}}_{test}$, where $\textbf{1}_q$ is an all ones vector. Given a training set $\mathcal{D}_{train}$ and a test set $\mathcal{D}_{test}$, the goal of PLC is to construct a weight matrix $\textbf{W}\in \mathbb{R}^{n\times n}$ representing the similarity of the examples for the dataset $\mathcal{D}=\mathcal{D}_{train}\cup\mathcal{D}_{test}$. For any example $\textbf{\textit{x}}_i$ in $\mathcal{D}$, we construct the k-nearest neighbors of $\textbf{\textit{x}}_i$, i.e $\mathcal{E}=\{(\textbf{\textit{x}}_i, \textbf{\textit{x}}_j)| \textbf{\textit{x}}_j\in KNN(\textbf{\textit{x}}_i),i\neq j \}$.

\subsection{Constructing Weight Matrix} Given a dataset $\mathcal{D}$ and its corresponding neighbor set $\mathcal{E}$, our objective is to derive a weight matrix $\textbf{W}={(w_{ij})}_{n\times n}$ that adequately represents the similarities among the examples. This has been proven to be effective in PLL and SC \cite{Zhang2015SolvingTP,10.5555/3504035.3504447}. Considering that higher weights should be assigned to examples with higher similarity, the weight matrix $\textbf{W}$ can be constructed by solving the following linear least square problem:
\begin{equation}
    \begin{aligned}
    &\mathop{\min}_{\textbf{W}}\sum_{j=1}^n\left\Vert\textbf{\textit{x}}_j-\sum\nolimits_{(\textbf{\textit{x}}_i, \textbf{\textit{x}}_j)\in \mathcal{E}}w_{ij}\textbf{\textit{x}}_i\right\Vert_2^2\\
    &\begin{array}{r@{\quad}r@{}l@{\quad}l}
         \mathrm{s.t.}& \textbf{W}^\top\textbf{1}_n=\textbf{1}_n,\textbf{0}_{n\times n}\leq\textbf{W}\leq\textbf{N},
    \end{array}  
    \label{eq1}
\end{aligned}
\end{equation}
where $\textbf{1}_n\in\mathbb{R}^n$ is an all ones vector and $\textbf{0}_{n\times n}\in\mathbb{R}^{n\times n}$ is an all zeros matrix. Moreover, $\textbf{N}\in\{0,1\}^{n\times n}$ is defined as: $n_{ij}=1$, if $(\textbf{\textit{x}}_i, \textbf{\textit{x}}_j)\in \mathcal{E}$ and $n_{ij}=0$ otherwise, which indicates that $w_{ij}\geq0$ $(\forall(\textbf{\textit{x}}_i, \textbf{\textit{x}}_j)\in \mathcal{E})$ and $w_{ij}=0$ $(\forall(\textbf{\textit{x}}_i, \textbf{\textit{x}}_j)\notin \mathcal{E})$. The first constraint $\textbf{W}^\top\textbf{1}_n=\textbf{1}_n$ normalizes the weight matrix $\textbf{W}$ and the second constraint $\textbf{0}_{n\times n}\leq\textbf{W}\leq\textbf{N}$ ensures the non-negativity of the weight matrix $\textbf{W}$, while simultaneously ensuring that the weight matrix has values only for the k-nearest neighbors. By solving Eq. \eqref{eq1}, we can obtain a weight matrix\footnote{Strictly, when applied to clustering, the matrix $\textbf{W}$ should be symmetric. To simplify the model, we ignore symmetry and will subsequently apply symmetry processing to $\textbf{W}$, i.e., $(\textbf{W}+\textbf{W}^\top)/2$.} containing the similarities of examples in the feature space.

\subsection{Label Disambiguation} Denote $\textbf{F}=[\textbf{\textit{f}}_1, \textbf{\textit{f}}_2,..., \textbf{\textit{f}}_n]^\top\in\mathbb{R}^{n\times q}$ as the label confidence matrix, where $f_{ij}$ represents the probability of the $j$-th label being the ground-truth label of sample $\textbf{\textit{x}}_i$. We initialize the label confidence matrix as follows: $\textbf{F}_{ij}=1/\sum_jy_{ij}$ if $y_{ij}=1$, otherwise $\textbf{F}_{ij}=0$. The similarity relationships of examples in the feature space should be consistent in the label space, i.e., if two samples are similar in the feature space, they are more likely to share the same ground-truth label. We leverage the weight matrix $\textbf{W}$ to disambiguate the label confidence matrix $\textbf{F}$, which can be obtained by solving the following problem:
\begin{equation}
    \begin{aligned}
    &\mathop{\min}_{\textbf{F}}\sum_{j=1}^n\left\Vert\textbf{\textit{f}}_j-\sum\nolimits_{(\textbf{\textit{x}}_i, \textbf{\textit{x}}_j)\in \mathcal{E}}w_{ij}\textbf{\textit{f}}_i\right\Vert_2^2\\
    &\begin{array}{r@{\quad}r@{}l@{\quad}l}
        \mathrm{s.t.}& \textbf{F}\textbf{1}_q=\textbf{1}_n,\textbf{0}_{n\times q}\leq\textbf{F}\leq\textbf{Y},
    \end{array}  
    \label{eq2}
\end{aligned}
\end{equation}
where $\textbf{1}_q\in\mathbb{R}^q$ is an all ones vector and $\textbf{0}_{n\times q}\in\mathbb{R}^{n\times q}$ is an all zeros matrix. After the label disambiguation process, we select the class with the highest label confidence of example $\textbf{\textit{x}}_i$ as the pseudo-label $y^*_i=\mathop{\arg\max}_{j\in \mathcal{Y}}\textbf{\textit{f}}_{ij}$.

\subsection{Pairwise Constraint Propagation} Based on the pseudo-labels obtained from label disambiguation, we denote the set of must-links (MLs) as $\mathcal{M}$ and the set of cannot-links (CLs) as $\mathcal{C}$. If two examples $\textbf{\textit{x}}_i$ and $\textbf{\textit{x}}_j$ have the same pseudo-label, i.e., $y^*_i=y^*_j$, then $(\textbf{\textit{x}}_i, \textbf{\textit{x}}_j)\in \mathcal{M}$, otherwise $(\textbf{\textit{x}}_i, \textbf{\textit{x}}_j)\in \mathcal{C}$. Then the indicator matrices of the MLs and CLs can be defined as:
\begin{equation}
    \begin{aligned}
     \textbf{M}_{ij} & = 
    \begin{cases}
        1, & \mbox{if}~(\textbf{\textit{x}}_i, \textbf{\textit{x}}_j)\in \mathcal{M}, \\
        0, & \mbox{otherwise},
    \end{cases}\\
    \textbf{C}_{ij} & = 
    \begin{cases} 
        1, & \mbox{if}~(\textbf{\textit{x}}_i, \textbf{\textit{x}}_j)\in \mathcal{C}, \\
        0, & \mbox{otherwise},
    \end{cases}
    \label{eq4}
\end{aligned}
\end{equation}
MLs and CLs essentially represent the relationships between sample classes, i.e., the similarity relationship and the dissimilarity relationship respectively. However, the initial similarity and dissimilarity relationships are imprecise. To address this issue, we propagate the similarity matrix $\textbf{S}\in \mathbb{R}^{n\times n}$ and the dissimilarity matrix $\textbf{D}\in \mathbb{R}^{n\times n}$ through the weight matrix $\textbf{W}$ constructed in the feature space, i.e., if two samples are similar in the feature space, their similarity and dissimilarity codings should also be similar. Moreover, the similarity matrix $\textbf{S}$ and the dissimilarity matrix $\textbf{D}$ naturally form an adversarial relationship, i.e., if two samples have high (low) similarity, their dissimilarity should be low (high). Combining propagation and adversarial relationship together, we have the following objective: 
\begin{equation}
    \begin{aligned}
    &\mathop{\min}_{\textbf{D},\textbf{S}}\parallel\textbf{D}\odot\textbf{S}\parallel_1 + \alpha \mathrm{Tr}(\textbf{D}\textbf{L}\textbf{D}^\top)+\beta \mathrm{Tr}(\textbf{S}\textbf{L}\textbf{S}^\top)\\
    &\begin{array}{r@{\quad}l@{}l@{\quad}l}
         \mathrm{s.t.} &\textbf{S}\geq\textbf{0}_{n\times n}, \textbf{D}\geq\textbf{0}_{n\times n},\\
         & \textbf{S}_{ij}=\textbf{M}_{ij}, \textbf{D}_{ij}=\textbf{C}_{ij}, \enspace\mathrm{if} \enspace(\textbf{\textit{x}}_i, \textbf{\textit{x}}_j)\in \mathcal{M}\cup\mathcal{C},\\
    \end{array}  
    \label{eq5}
\end{aligned}
\end{equation}
where $\odot$ represents the elementwise product and $||\cdot||_1$ refers to the $l_1$ norm, i.e., $||\textbf{D}\odot\textbf{S}||_1=\sum_{ij}\textbf{D}_{ij}\textbf{S}_{ij}$. $\textbf{L}\in \mathbb{R}^{n\times n}$ is the Laplacian matrix of weight matrix $\textbf{W}$, i.e., $\textbf{L}=\textbf{A}-\textbf{W}$, where $\textbf{A}$ is a diagonal matrix with the $i$-th diagonal element $\textbf{A}_{ii}=\sum_{j=1}^n \textbf{W}_{ij}$. $\alpha, \beta\geq0$ are two hype-parameters to balance different terms. Through Eq. (\ref{eq5}), we can enhance the accuracy and density of information in $\mathbf{S}$ and $\mathbf{D}$ by incorporating adversarial and propagation terms.

\begin{algorithm}[tb]
    \caption{Pseudo-code of PLC}
    \label{alg:algorithm}
    \textbf{Input}: \\$\mathcal{D}$ :\quad the partial label training dataset $\{(\textit{\textbf{x}}_i,\textbf{Y}_i)|1\leq i\leq n\}$.\\
    $k$ :\quad the number of nearest neighbors.\\
    $l$ :\quad the number of clusters.\\
    $\alpha$, $\beta$, $\gamma$ :\quad the hyper-parameters in Eq.\eqref{eq6} and Eq.\eqref{eq10}.\\
    $\textit{\textbf{x}}^*$ : the unseen test example for prediction.\\
    \textbf{Output}:\\ $\textit{\textbf{C}}$ :\quad the clusters $\{C_1,C_2,...,C_l\}.$
    \begin{algorithmic}[1] 
        \STATE Construct partial label dataset of all samples $\mathcal{D}^{'} =\mathcal{D}\bigcup\{(\textit{\textbf{x}}^*,\textbf{Y}^*_i)|\textbf{Y}^*_i=\textbf{1}_q\}$ ;
        \STATE Initialize the weight matrix $\textbf{W}$ by solving Eq. \eqref{eq1};
        \STATE Initialize the label confidence matrix $\textbf{F}$ by solving Eq. \eqref{eq2};
        \STATE Initialize pairwise constraints according to Eq. \eqref{eq4};
        \STATE Initialize the similarity matrix $\textbf{S}$ and dissimilarity matrix $\textbf{D}$ by solving Eq. \eqref{eq5};
        \REPEAT
        \FOR{$j=1$ $\textbf{to}$ $n$} 
        \STATE  Update the vector $\hat{\textbf{\textit{w}}}_j\in \mathbb{R}^k$ by solving Eq. \eqref{eq8};
        \ENDFOR
        \STATE Update the label confidence matrix $\textbf{F}$ by solving Eq. \eqref{eq3} or iteratively solving the subproblems as Eq. \eqref{eq9};
        \STATE Update the matrix $\textbf{S}$ and $\textbf{D}$ by solving Eq. \eqref{eq10} with KKT conditions;
        \UNTIL convergence or maximum number of iterations being reached.
        \STATE Perform SC on $\textbf{W}$ to obtain the final clustering results $\textbf{C}$.\STATE \textbf{return} the final clustering results $\textit{\textbf{C}}=\{C_1,C_2,...,C_l\}.$
    \end{algorithmic}
\end{algorithm}

\subsection{Overall Model} Combining the optimization problems mentioned above, the final optimization objective of our proposed PLC model can be summarized as follows:
\begin{small}
    \begin{align}
    &\mathop{\min}_{\textbf{W},\textbf{F},\textbf{D},\textbf{S}}\sum_{j=1}^n\left\Vert\textbf{\textit{x}}_j-\sum\nolimits_{(\textbf{\textit{x}}_i, \textbf{\textit{x}}_j)\in \mathcal{E}}w_{ij}\textbf{\textit{x}}_i\right\Vert_2^2+\alpha \mathrm{Tr}(\textbf{D}\textbf{L}\textbf{D}^\top)\nonumber\\
    &+\sum_{j=1}^n\left\Vert\textbf{\textit{f}}_j-\sum\nolimits_{(\textbf{\textit{x}}_i, \textbf{\textit{x}}_j)\in \mathcal{E}}w_{ij}\textbf{\textit{f}}_i\right\Vert_2^2+\beta \mathrm{Tr}(\textbf{S}\textbf{L}\textbf{S}^\top)+\parallel\textbf{D}\odot\textbf{S}\parallel_1\nonumber\\
    &\begin{array}{l@{\quad}l@{}l@{\quad}l@{\quad}}
         \mathrm{s.t.} &\textbf{W}^\top\textbf{1}_n=\textbf{1}_n,\textbf{0}_{n\times n}\leq\textbf{W}\leq\textbf{N},\\
        &\textbf{F}\textbf{1}_q=\textbf{1}_n,\textbf{0}_{n\times q}\leq\textbf{F}\leq\textbf{Y},\\ &\textbf{S}\geq\textbf{0}_{n\times n}, \textbf{D}\geq\textbf{0}_{n\times n},\\
         & \textbf{S}_{ij}=\textbf{M}_{ij}, \textbf{D}_{ij}=\textbf{C}_{ij}, \enspace\mathrm{if} \enspace(\textbf{\textit{x}}_i, \textbf{\textit{x}}_j)\in \mathcal{M}\cup\mathcal{C}.\\
    \end{array} 
    \label{eq6}
\end{align}
\end{small}

\subsection{Alternating Optimization} Since the optimization objective Eq. \eqref{eq6} consists of multiple linear and quadratic equations and has multiple optimization variables, it is difficult to optimize directly. In this subsection, we solve this problem through alternating optimization methods, i.e., alternately optimizing one variable and fixing the other variables until convergence or reaching the predetermined maximum iterations.

\begin{figure*}[t]
    \centering
    \subfigure{
    \includegraphics[width=3.5cm,height=3.5cm]{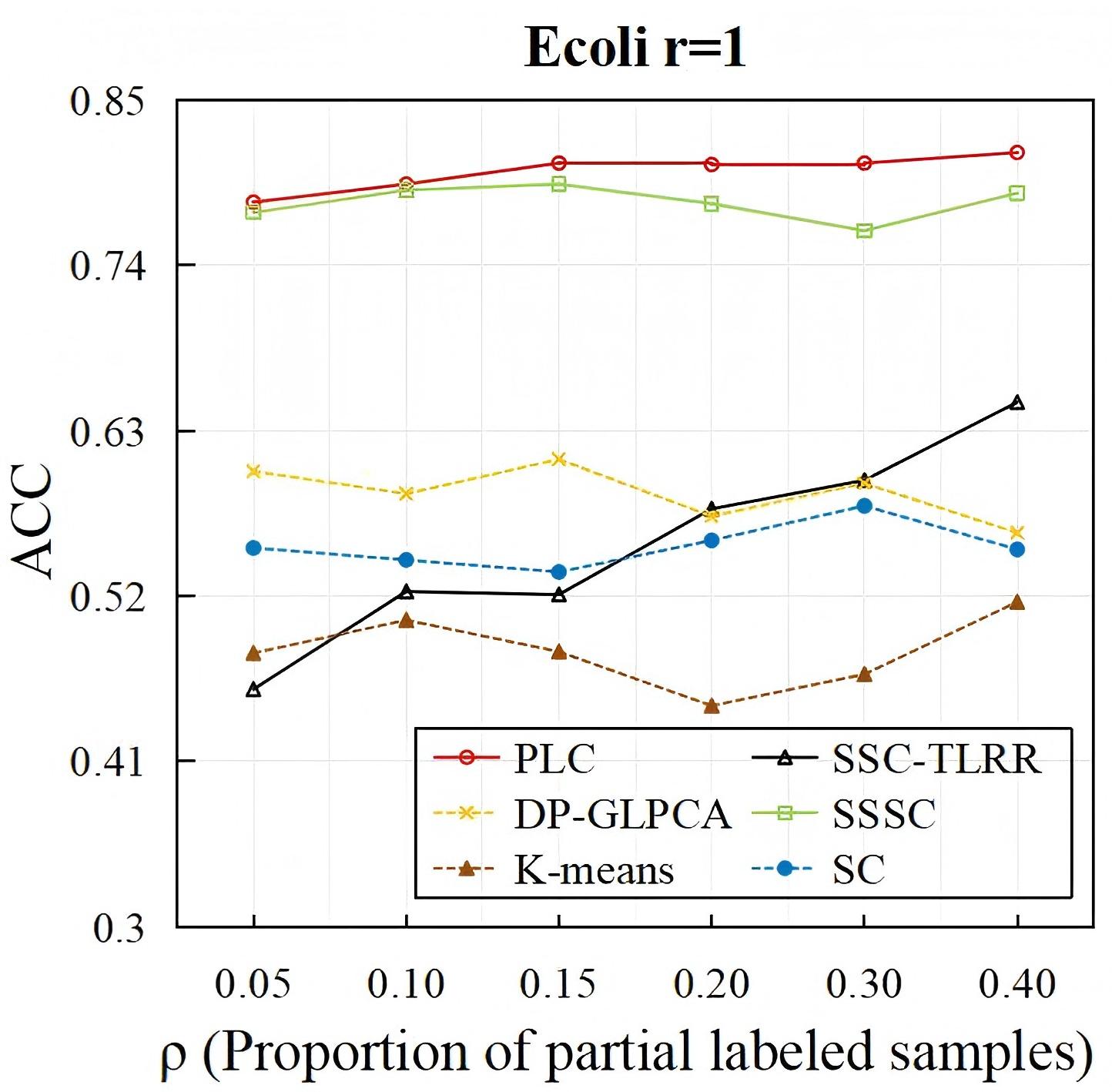} }
    \hspace{2.5mm}
    \subfigure{
    \includegraphics[width=3.5cm,height=3.5cm]{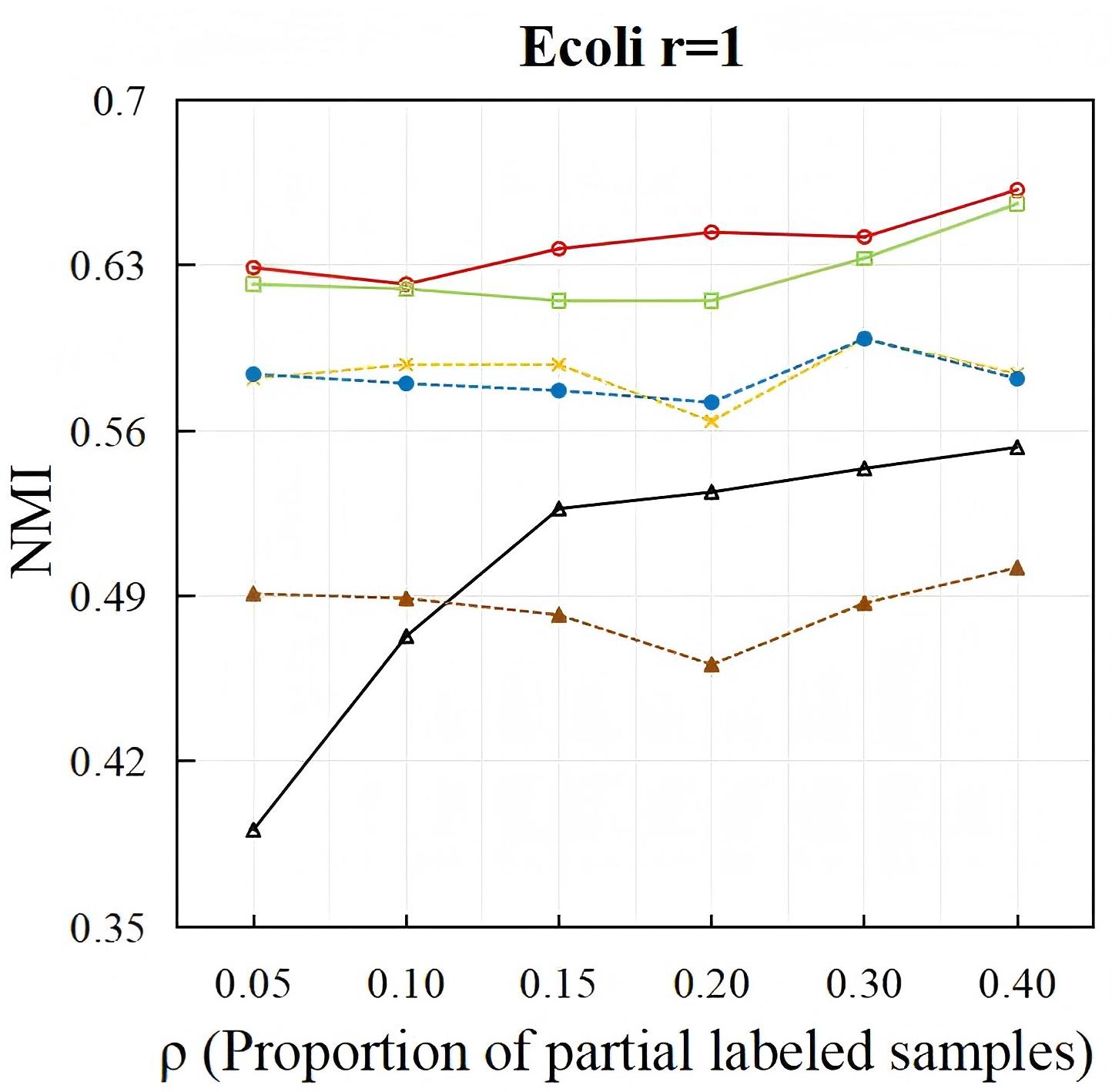} }	
    \hspace{2.5mm}
    \subfigure{
    \includegraphics[width=3.5cm,height=3.5cm]{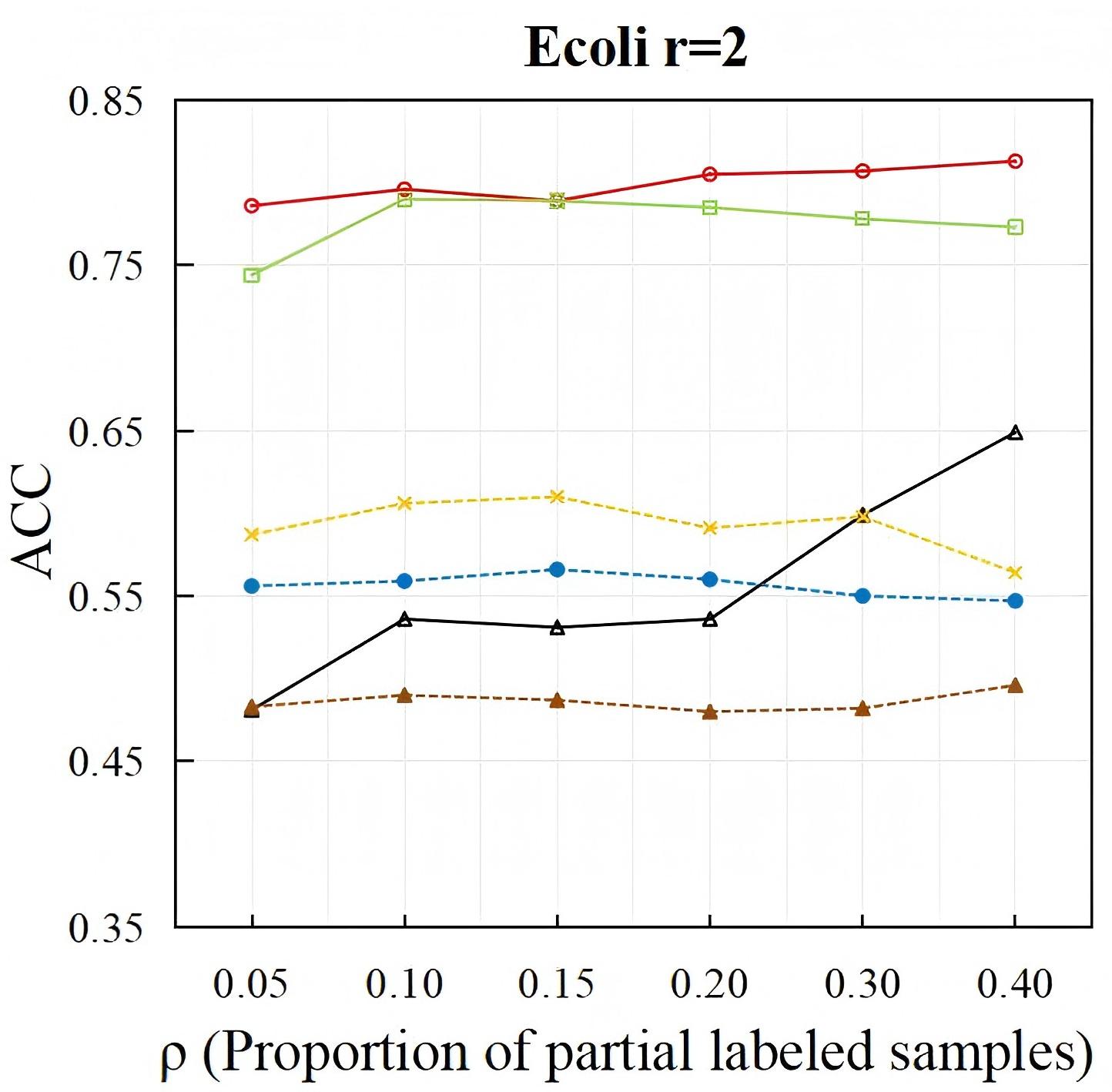} }
    \hspace{2.5mm}
    \subfigure{
    \includegraphics[width=3.5cm,height=3.5cm]{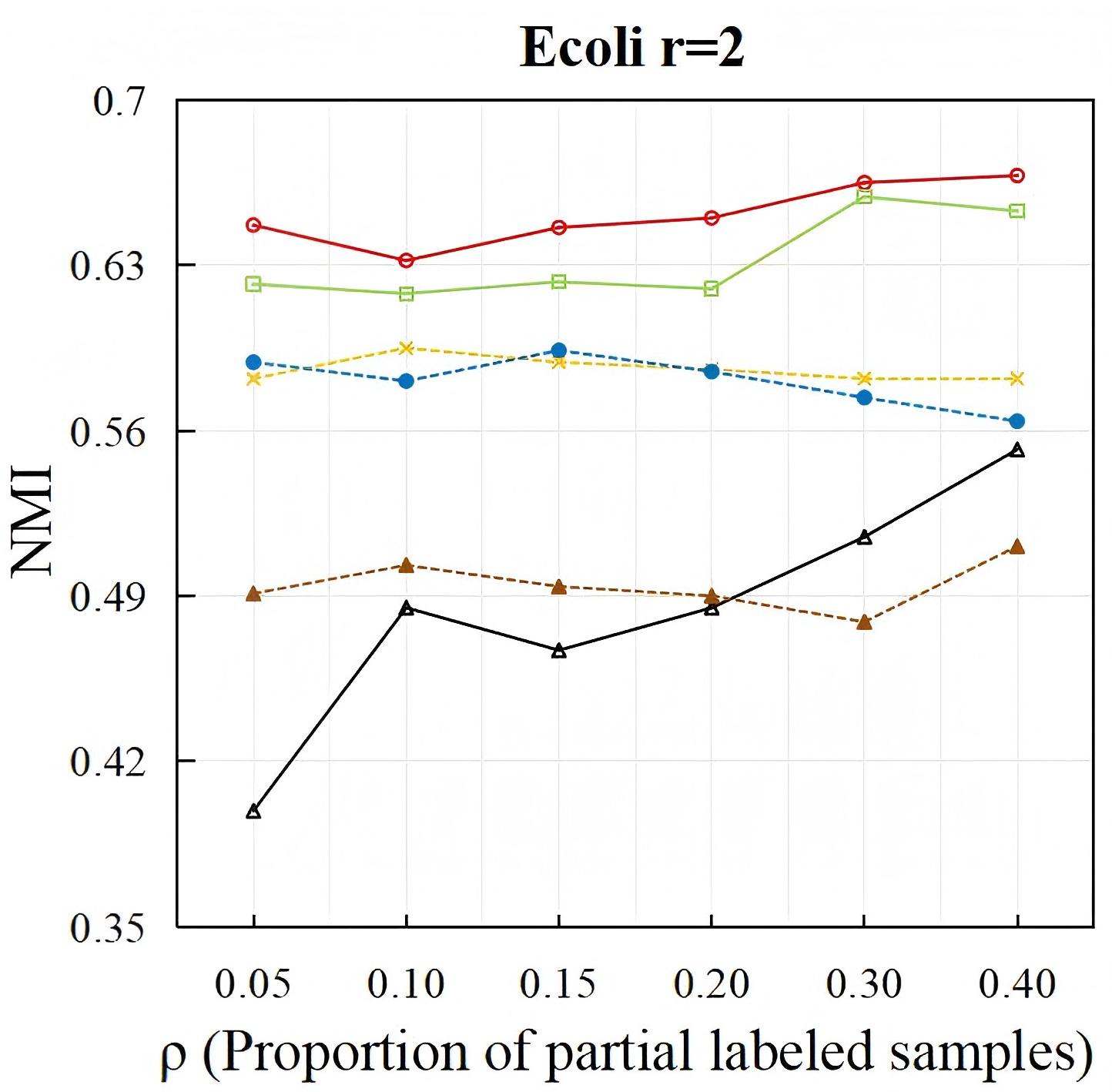} }
    \hspace{2.5mm}

    \centering
    \subfigure{
    \includegraphics[width=3.5cm,height=3.5cm]{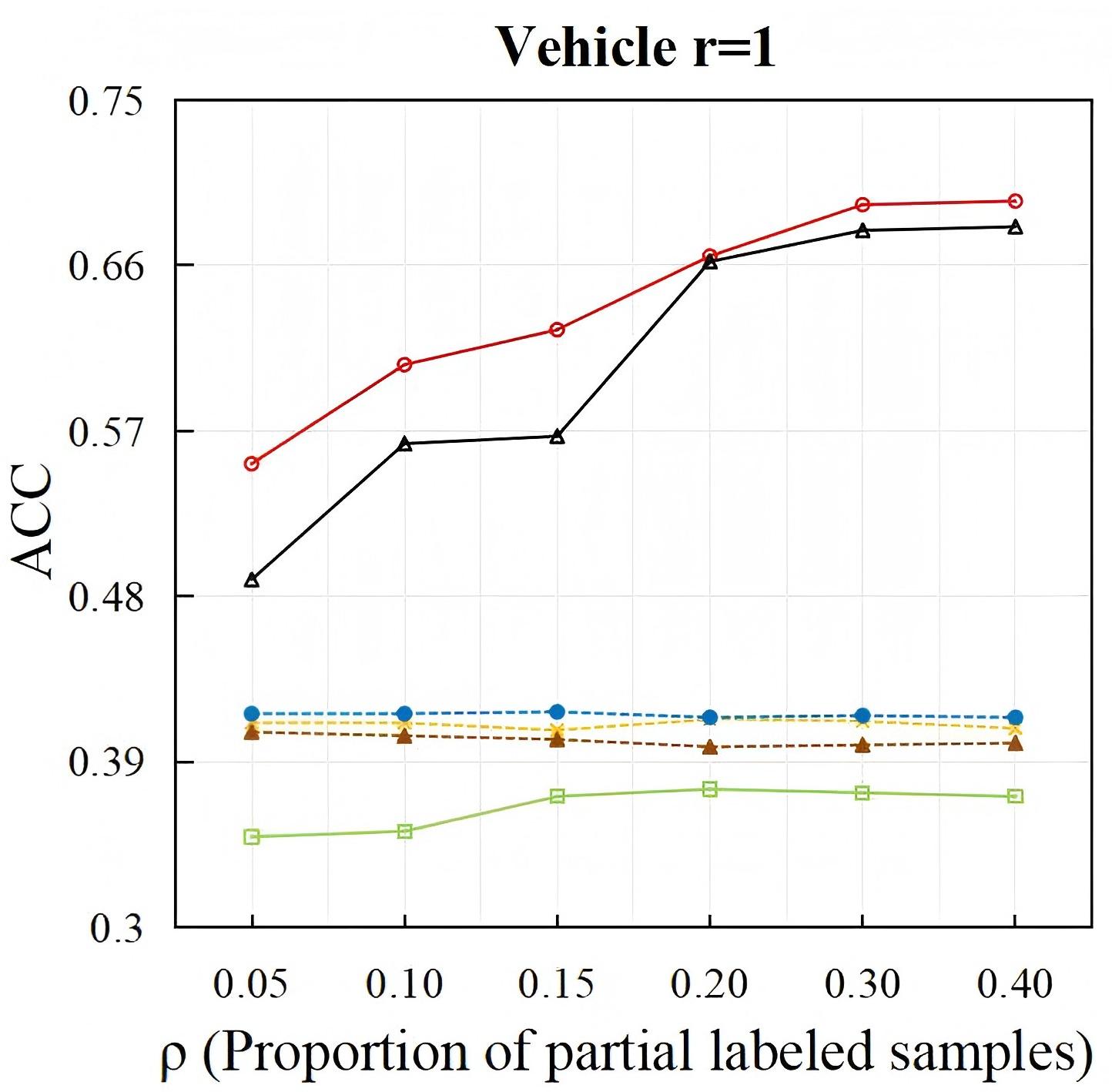} }
    \hspace{2.5mm}
    \subfigure{
    \includegraphics[width=3.5cm,height=3.5cm]{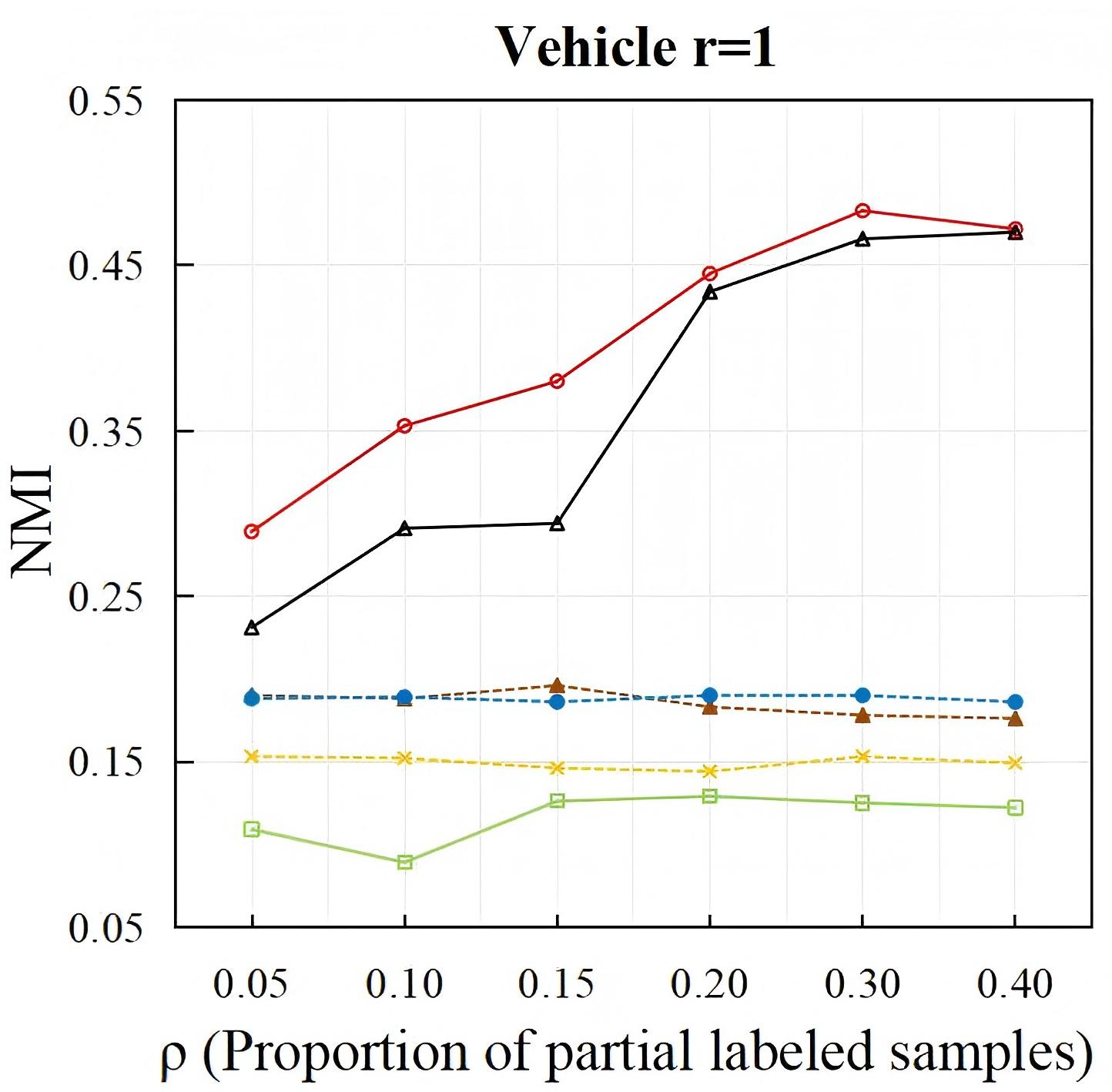} }	
    \hspace{2.5mm}
    \subfigure{
    \includegraphics[width=3.5cm,height=3.5cm]{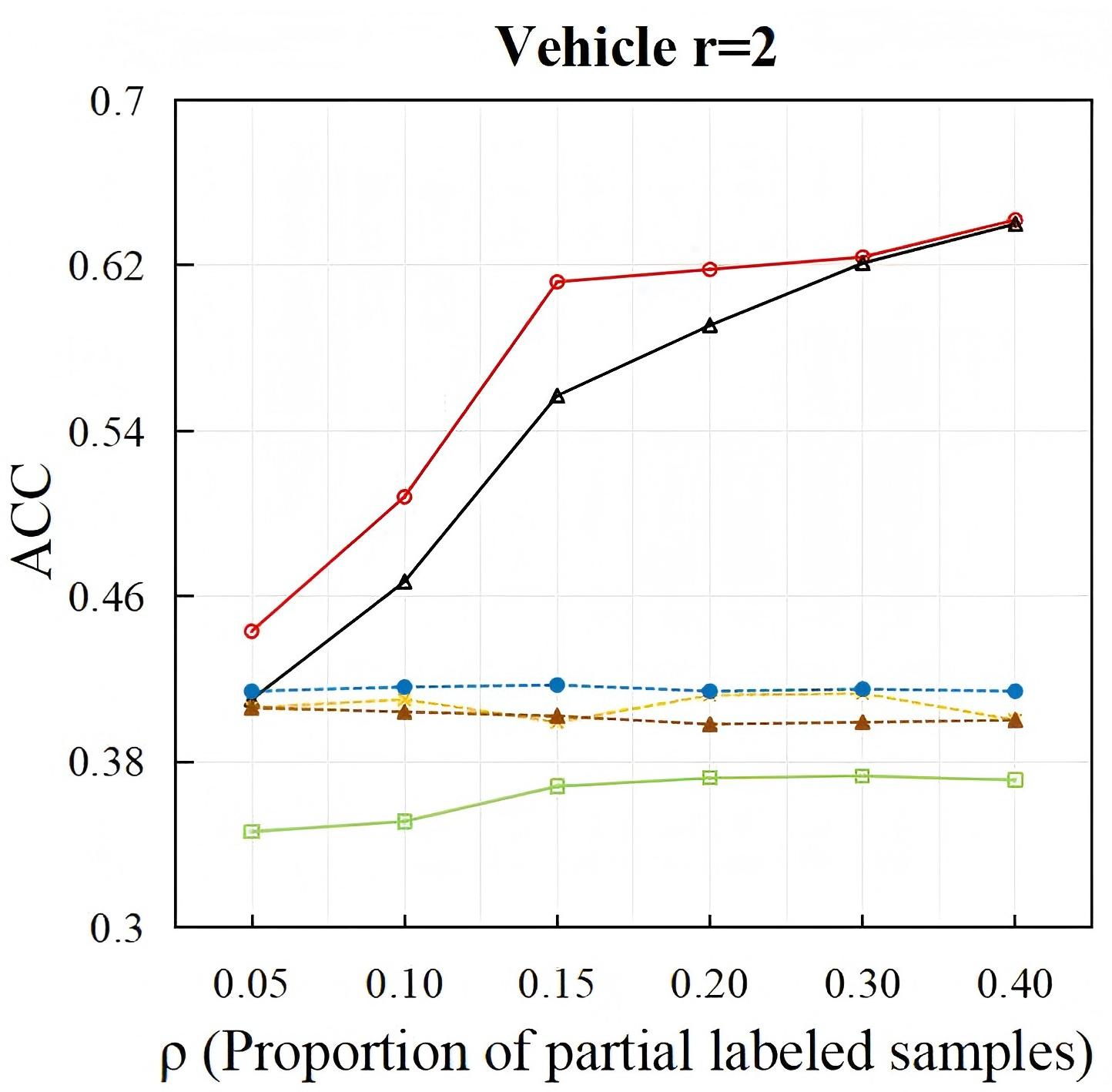} }
    \hspace{2.5mm}
    \subfigure{
    \includegraphics[width=3.5cm,height=3.5cm]{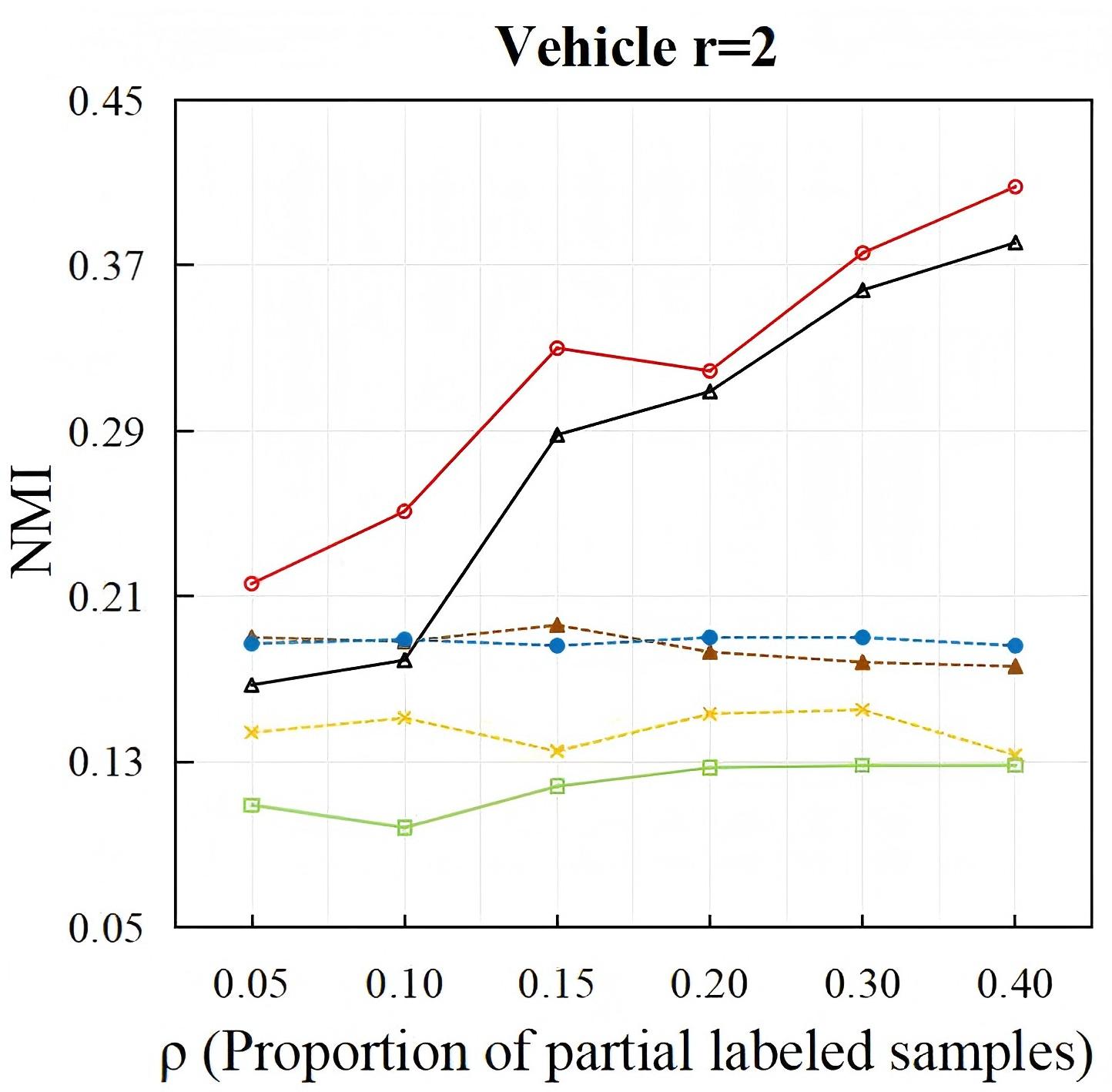} }
    \hspace{2.5mm}

    \centering
    \subfigure{
    \includegraphics[width=3.5cm,height=3.5cm]{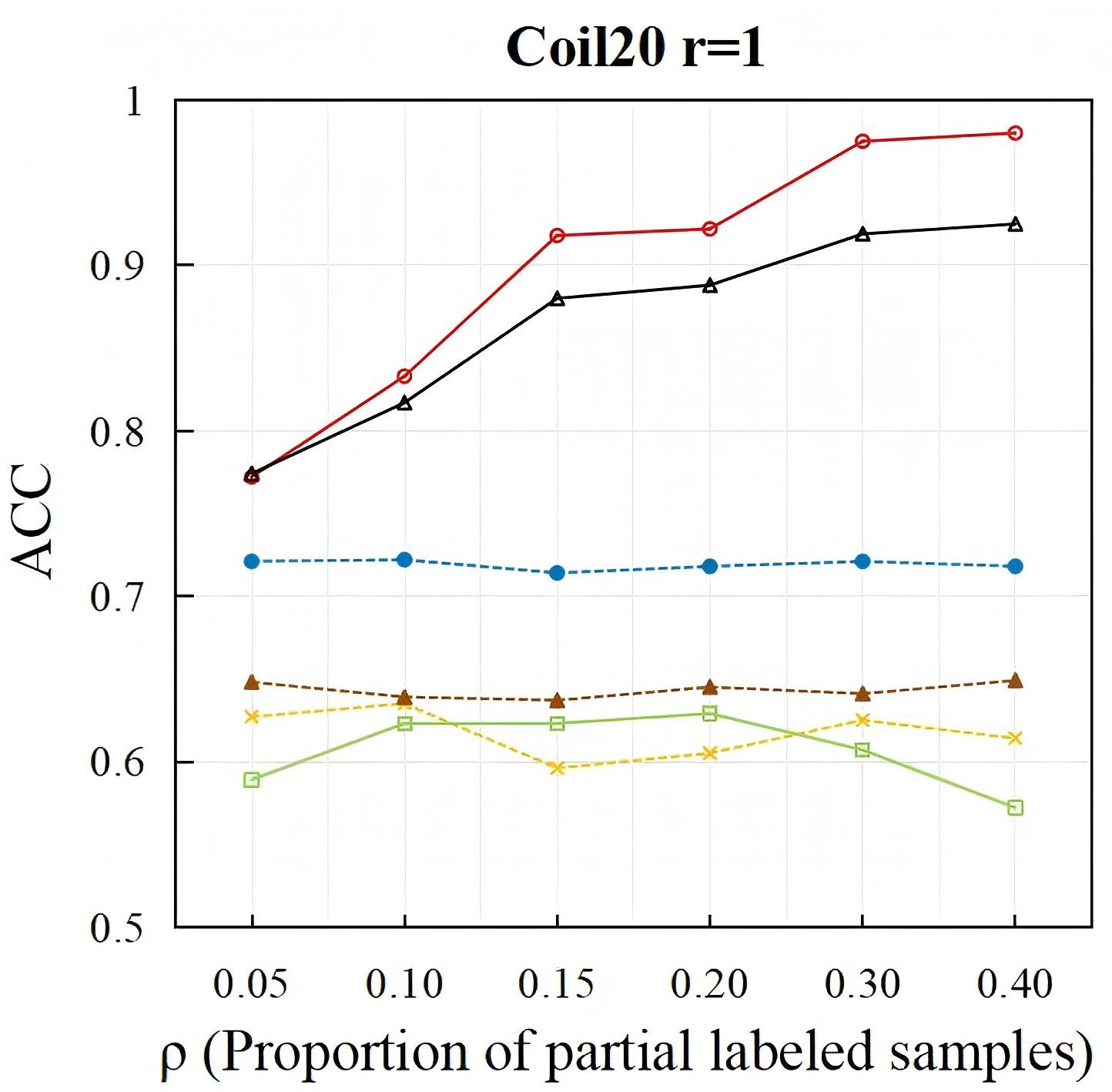} }
    \hspace{2.5mm}
    \subfigure{
    \includegraphics[width=3.5cm,height=3.5cm]{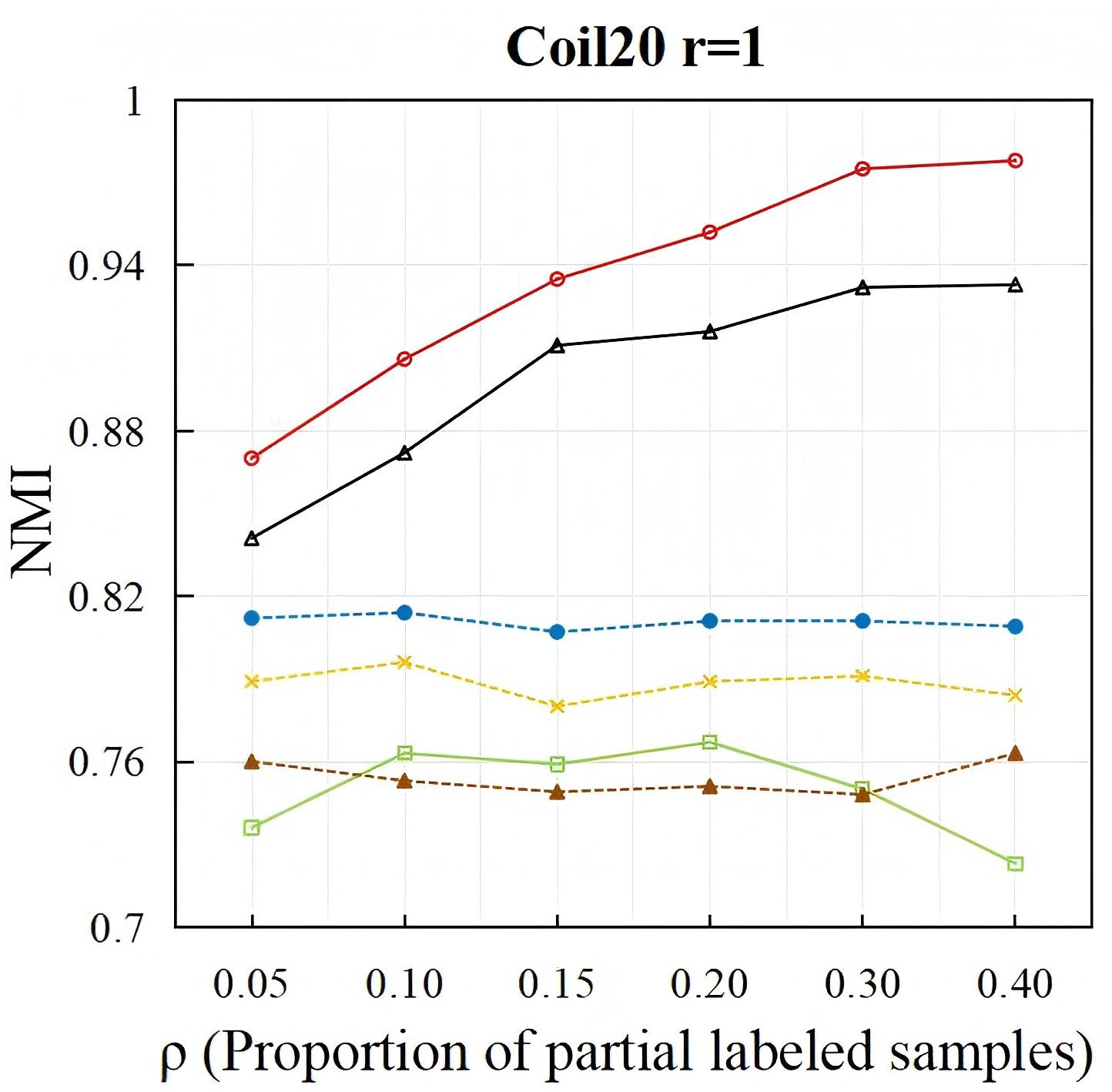} }	
    \hspace{2.5mm}
    \subfigure{
    \includegraphics[width=3.5cm,height=3.5cm]{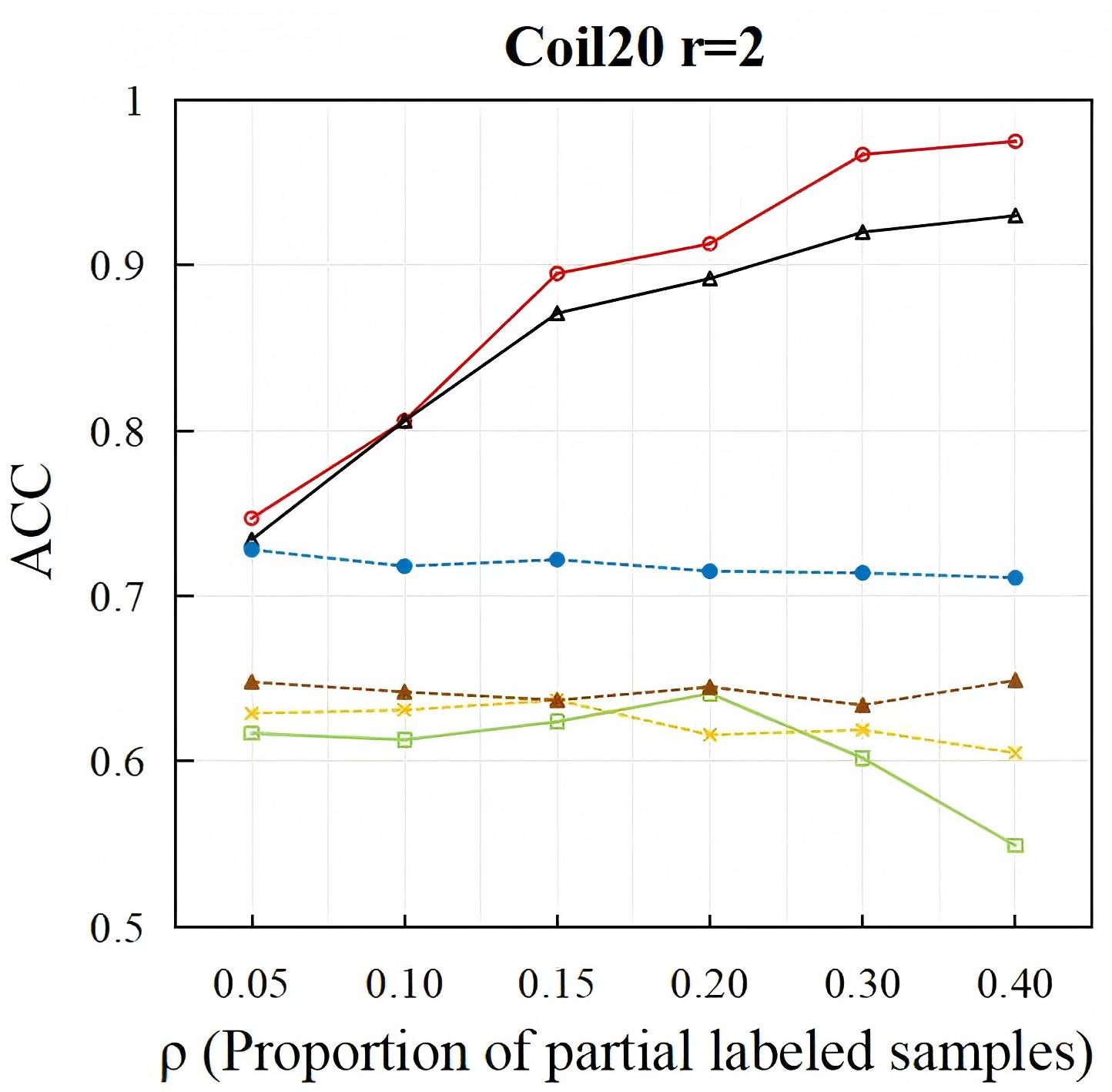} }
    \hspace{2.5mm}
    \subfigure{
    \includegraphics[width=3.5cm,height=3.5cm]{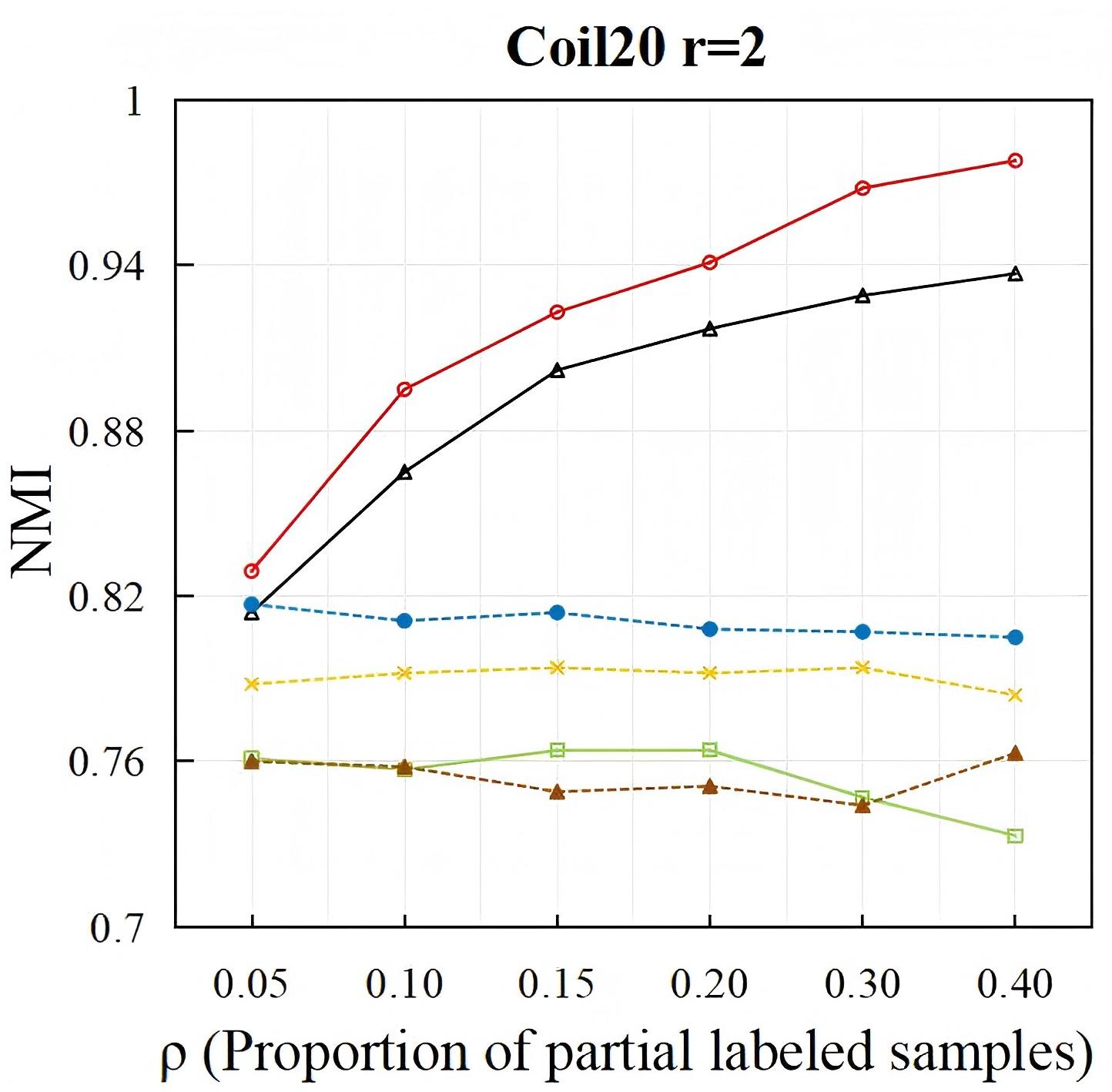} }
    \hspace{2.5mm}
    
    \caption{ ACCs and NMIs when compared with constrained clustering methods under different proportions of partial label training examples on synthetic UCI datasets. }
    \label{f1}
\end{figure*}

\begin{table*}[t]
\small
\renewcommand{\arraystretch}{0.9}
\centering
\resizebox{1\textwidth}{!}{\begin{tabular}{ccccccc}
\Xhline{.5px}
\Xhline{.5px} 
\multicolumn{1}{l}{\multirow{2}{2cm}{\centering Compared \\ Method}} & \multicolumn{6}{c}{\textbf{Lost}}                   \\ \cline{2-7} 
\multicolumn{1}{l}{}  & $\rho=0.05$ & $\rho=0.10$ & $\rho=0.15$ & $\rho=0.20$ & $\rho=0.30$ & $\rho=0.40$  \\ \hline
\textbf{PLC (Ours)}      & $\mathbf{0.399\pm0.027}$       & $\mathbf{0.497\pm0.024}$  & $\mathbf{0.512\pm0.024}$       & $\mathbf{0.553\pm0.026}$    & $\mathbf{0.608\pm0.021}$       & $\mathbf{0.641\pm0.022}$      \\
\textbf{K-means}     & $0.205\pm0.013$       & $0.204\pm0.011$      & $0.222\pm0.013$      & $0.203\pm0.013$  &$0.202\pm0.010$  & $0.209\pm0.017$            \\
\textbf{SC}     & $0.285\pm0.019$       & $\underline{0.309\pm0.027}$      & $\underline{0.305\pm0.012}$      & $\underline{0.306\pm0.011}$  &$0.317\pm0.018$  & $0.303\pm0.021$            \\
\textbf{SSC-TLRR}    & $0.266\pm0.029$       & $0.233\pm0.014$       & $0.243\pm0.010$       & $0.273\pm0.014$   &$\underline{0.328\pm0.008}$ &$\underline{0.397\pm0.015}$          \\
\textbf{DP-GLPCA}   & $0.208\pm0.018$       & $0.202\pm0.024$       & $0.211\pm0.027$       & $0.220\pm0.014$ &$0.200\pm0.018$ &$0.209\pm0.026$           \\
\textbf{SSSC}   & $\underline{0.296\pm0.027}$       & $0.299\pm0.015$       & $0.304\pm0.027$       & $0.304\pm0.021$ &$0.298\pm0.020$ & $0.304\pm0.018$          \\ \Xhline{.5px}
\Xhline{.5px} 
\multicolumn{1}{l}{\multirow{2}{2cm}{\centering Compared \\ Method}} & \multicolumn{6}{c}{\textbf{MSRCv2}}                   \\ \cline{2-7} 
\multicolumn{1}{l}{}  & $\rho=0.05$ & $\rho=0.10$ & $\rho=0.15$ & $\rho=0.20$ &$\rho=0.30$ &$\rho=0.40$\\ \hline
\textbf{PLC (Ours)}      & $\mathbf{0.337\pm 0.026}$      &$\mathbf{0.359\pm0.020}$        & $\mathbf{0.372\pm0.021}$       &  $\mathbf{0.404\pm0.016}$  &$\mathbf{0.425\pm0.015}$ &$\mathbf{0.433\pm0.017}$     \\
\textbf{K-means}     &  $0.243\pm0.022$   &$0.229\pm0.008$         & $0.245\pm0.022$      & $0.266\pm0.017$  &$0.235\pm0.015$ &$0.265\pm0.015$              \\
\textbf{SC}     &  $0.297\pm0.011$      &$0.293\pm0.006$        & $0.293\pm0.010$       &$0.295\pm0.012$  &$0.284\pm0.010$   & $0.291\pm0.008$             \\
\textbf{SSC-TLRR}    & $0.298\pm0.013$       & $0.263\pm0.012$       & $0.261\pm0.010$       & $0.269\pm0.010$    &$0.301\pm0.018$ &$\underline{0.355\pm0.013}$          \\
\textbf{DP-GLPCA}   & $\underline{0.329\pm0.016}$       & $0.331\pm0.011$       & $0.314\pm0.018$       & $\underline{0.350\pm0.021}$ &$\underline{0.321\pm0.014}$ &$0.317\pm0.014$          \\
\textbf{SSSC}   & $0.297\pm0.015$       & $\underline{0.337\pm0.023}$       & $\underline{0.354\pm0.012}$       & $0.337\pm0.026$ &$0.294\pm0.023$  &$0.280\pm0.016$           \\  \Xhline{.5px}
\Xhline{.5px} 
\multicolumn{1}{l}{\multirow{2}{2cm}{\centering Compared \\ Method}} & \multicolumn{6}{c}{\textbf{Mirflickr}}                   \\ \cline{2-7} 
\multicolumn{1}{l}{}  & $\rho=0.05$ & $\rho=0.10$ & $\rho=0.15$ & $\rho=0.20$ &$\rho=0.30$ &$\rho=0.40$ \\ \hline
\textbf{PLC (Ours)}  & $\mathbf{0.567\pm0.021}$       &  $\mathbf{0.585\pm0.014}$    &$\mathbf{0.579\pm0.020}$ &$\mathbf{0.587\pm0.022}$ &$\mathbf{0.589\pm0.022}$  &$\mathbf{0.596\pm0.022}$        \\
\textbf{K-means}     & $0.299\pm0.008$       & $0.296\pm0.007$       & $0.299\pm0.017$       &  $0.304\pm0.008$  &$0.298\pm0.008$       & $0.303\pm0.015$         \\
\textbf{SC}     & $0.282\pm0.006$        &$0.283\pm0.001$         & $0.283\pm0.003$        &$0.285\pm0.002$  &$0.284\pm0.004$      &  $0.284\pm0.003$              \\
\textbf{SSC-TLRR}    & $0.415\pm0.022$       & $0.423\pm0.021$       & $0.359\pm0.018$       &$0.385\pm0.042$    &$0.435\pm0.010$ & $\underline{0.474\pm0.012}$         \\
\textbf{DP-GLPCA}   & $\underline{0.465\pm0.013}$       & $\underline{0.460\pm0.012}$       & $\underline{0.460\pm0.003}$       & $\underline{0.464\pm0.004}$     & $\underline{0.465\pm0.013}$    & $0.459\pm0.005$   \\
\textbf{SSSC}   & $0.352\pm0.002$       & $0.350\pm0.004$       & $0.351\pm0.014$       & $0.363\pm0.021$  & $0.364\pm0.012$ & $0.366\pm0.020$          \\  \Xhline{.5px}
\Xhline{.5px}
\multicolumn{1}{l}{\multirow{2}{2cm}{\centering Compared \\ Method}} & \multicolumn{6}{c}{\textbf{BirdSong}}                   \\ \cline{2-7} 
\multicolumn{1}{l}{}  & $\rho=0.05$ & $\rho=0.10$ &$\rho=0.15$ & $\rho=0.20$  &$\rho=0.30$ &$\rho=0.40$\\ \hline
\textbf{PLC (Ours)}      & $\mathbf{0.612\pm 0.018}$      &$\mathbf{0.627\pm0.020}$        & $\mathbf{0.632\pm0.024}$       &  $\mathbf{0.638\pm0.031}$  &  $\mathbf{0.643\pm0.027}$   &  $\mathbf{0.654\pm0.023}$         \\
\textbf{K-means}     & $0.280\pm0.041$       & $0.303\pm0.022$       & $0.283\pm0.031$       &  $0.285\pm0.026$  &$0.286\pm0.028$       & $0.281\pm0.029$             \\
\textbf{SC}     & $0.437\pm0.003$     &  $0.437\pm0.004$      &  $0.437\pm0.004$      &  $0.442\pm0.004$  & $0.440\pm0.004$     &  $0.440\pm0.003$           \\
\textbf{SSC-TLRR}    & $0.403\pm0.016$       & $0.431\pm0.042$       & $0.480\pm0.045$          &$\underline{0.491\pm0.038}$ &$\underline{0.569\pm0.012}$ &$\underline{0.590\pm0.016}$         \\
\textbf{DP-GLPCA}   & $\underline{0.476\pm0.002}$       & $\underline{0.482\pm0.003}$       & $\underline{0.485\pm0.003}$       & $0.488\pm0.007$     & $0.494\pm0.007$    & $0.490\pm0.006$   \\
\textbf{SSSC}   & $0.453\pm0.013$       & $0.471\pm0.016$       & $0.479\pm0.020$       & $0.484\pm0.023$  & $0.442\pm0.032$ & $0.423\pm0.017$          \\   \Xhline{.5px}
\Xhline{.5px} 
\end{tabular}}
\caption{Experimental results on ACC when compared with constrained clustering methods under different proportions of partial label training examples on real-world datasets, where bold and underlined indicate the best and
second best results respectively.}
\label{tb2}
\end{table*}

\noindent\textbf{Update $\textbf{W}$} \enspace With fixed $\textbf{F}$, $\textbf{S}$ and $\textbf{D}$, the subproblem of variable $\textbf{W}$ can be formulated as:
\begin{equation}
    \begin{aligned}
    &\mathop{\min}_{\textbf{W}}\sum_{j=1}^n\left\Vert\textbf{\textit{x}}_j-\sum\nolimits_{(\textbf{\textit{x}}_i, \textbf{\textit{x}}_j)\in \mathcal{E}}w_{ij}\textbf{\textit{x}}_i\right\Vert_2^2+\alpha \mathrm{Tr}(\textbf{D}\textbf{L}\textbf{D}^\top)\\ 
    &+\sum_{j=1}^n\left\Vert\textbf{\textit{f}}_j-\sum\nolimits_{(\textbf{\textit{x}}_i, \textbf{\textit{x}}_j)\in \mathcal{E}}w_{ij}\textbf{\textit{f}}_i\right\Vert_2^2+\beta \mathrm{Tr}(\textbf{S}\textbf{L}\textbf{S}^\top)\\
    &\begin{array}{l@{\quad}l@{}l@{\quad}l@{\quad}}
         \mathrm{s.t.} &\textbf{W}^\top\textbf{1}_n=\textbf{1}_n,\textbf{0}_{n\times n}\leq\textbf{W}\leq\textbf{N}.\\
    \end{array} 
    \label{eq7}
\end{aligned}
\end{equation}
Eq. \eqref{eq7} can be separated column-wisely and we can optimize the $j$-th column vector $\textbf{W}_{\cdot j}$ while fixing other column vectors. For the vector $\textbf{W}_{\cdot j}$, only $k$ non-negative elements located in the neighborhood of the sample $\textbf{\textit{x}}_j$ need to be updated. We can simplify the optimization vector $\textbf{W}_{\cdot j}$ as $\hat{\textbf{\textit{w}}}_j\in \mathbb{R}^k$. Denote the matrix $\textbf{O}^{\textit{f}_j}=[\textbf{\textit{f}}_j-\textbf{\textit{f}}_{\mathcal{N}_j(1)}, \textbf{\textit{f}}_j-\textbf{\textit{f}}_{\mathcal{N}_j(2)},..., \textbf{\textit{f}}_j-\textbf{\textit{f}}_{\mathcal{N}_j(k)}]^\top\in \mathbb{R}^{k\times q}$ and the matrix $\textbf{O}^{\textit{x}_j}=[\textbf{\textit{x}}_j-\textbf{\textit{x}}_{\mathcal{N}_j(1)}, \textbf{\textit{x}}_j-\textbf{\textit{x}}_{\mathcal{N}_j(2)},..., \textbf{\textit{x}}_j-\textbf{\textit{x}}_{\mathcal{N}_j(k)}]^\top\in \mathbb{R}^{k\times d}$, where $\mathcal{N}_j(\cdot)$ represents the $k$ nearest neighbors of the sample $\textbf{\textit{x}}_j$. The subproblem of Eq. \eqref{eq7} can be reformulated as:
\begin{equation}
    \begin{aligned}
    &\mathop{\min}_{\hat{\textbf{\textit{w}}}_j}\hat{\textbf{\textit{w}}}_j^\top(\textbf{G}^{\textit{x}_j}+\textbf{G}^{\textit{f}_j})\hat{\textbf{\textit{w}}}_j+(\alpha\textbf{H}_{\cdot j}^\top+\beta\textbf{K}_{\cdot j}^\top)\hat{\textbf{\textit{w}}}_j\\
    &\begin{array}{l@{\quad}l@{}l@{\quad}l@{\quad}}
         \mathrm{s.t.} &\hat{\textbf{\textit{w}}}_j^\top\textbf{1}_k=1, \textbf{0}_{k}\leq\hat{\textbf{\textit{w}}}_j\leq\textbf{1}_{k},\\
    \end{array} 
    \label{eq8}
\end{aligned}
\end{equation}
where $\textbf{G}^{\textit{x}_j}=\textbf{O}^{\textit{x}_j}(\textbf{O}^{\textit{x}_j})^\top\in\mathbb{R}^{k\times k}\mathrm{,} \ \textbf{G}^{\textit{f}_j}=\textbf{O}^{\textit{f}_j}(\textbf{O}^{\textit{f}_j})^\top\in \mathbb{R}^{k\times k}$ are two Gram matrices for feature vector $\textbf{\textit{x}}_j$ and label confidence vector $\textbf{\textit{f}}_j$ respectively. $\textbf{H}_{\cdot j}=[H_{1j},H_{2j},...,H_{kj}]^\top\in \mathbb{R}^{k\times 1}$, where $H_{ij}=\parallel\textbf{D}_{\cdot \mathcal{N}_j(i)}-\textbf{D}_{\cdot j}\parallel^2_2$ represents the difference in dissimilarity encoding between the sample $\textit{x}_j$ and the neighboring sample $\textit{x}_i$. $\textbf{K}_{\cdot j}=[K_{1j},K_{2j},...,K_{kj}]^\top\in \mathbb{R}^{k\times 1}$, where $K_{ij}=\parallel\textbf{S}_{\cdot \mathcal{N}_j(i)}-\textbf{S}_{\cdot j}\parallel^2_2$ represents the difference in similarity encoding between the sample $\textit{x}_j$ and the neighboring sample $\textit{x}_i$. Obviously, Eq. \eqref{eq8} is a standard Quadratic Programming (QP) problem and it can be efficiently solved by any QP tools.

\noindent\textbf{Update $\textbf{F}$}\enspace With fixed $\textbf{W}$, $\textbf{S}$ and $\textbf{D}$, the subproblem of variable $\textbf{F}$ is the same as Eq. \eqref{eq2}. Note that $\sum\nolimits_{(\textbf{\textit{x}}_i, \textbf{\textit{x}}_j)\in \mathcal{E}}w_{ij}=1$, we can define $\textbf{T}=\textbf{W}\textbf{W}^\top+\textbf{I}_{n\times n}\textbf{W}^\top\textbf{1}_{n\times n}\textbf{W}-2\textbf{W}$, where $\textbf{I}_{n\times n}$ is an $n$-order identity matrix and $\textbf{1}_{n\times n}$ is an matrix of all ones. The Eq. \eqref{eq2} can be rewritten as:
\begin{equation}
    \begin{aligned}
    &\mathop{\min}_{\textbf{F}}\sum_{i=1}^n\sum_{j=1}^nt_{ij}\textbf{\textit{f}}_i^\top \textbf{\textit{f}}_j\\
    &\begin{array}{r@{\quad}r@{}l@{\quad}l}
        \mathrm{s.t.}& \textbf{F}\textbf{1}_q=\textbf{1}_n,\textbf{0}_{n\times q}\leq\textbf{F}\leq\textbf{Y}.
    \end{array}  
    \label{eq3}
\end{aligned}
\end{equation}
This is also a QP problem, which can be solved by any QP tools. However, this QP problem contains $nq$ variables and $n(q+1)$ constraints, which leads to excessive computational overhead when $nq$ is large. According to \cite{inproceedings}, we can update the label confidence vector $\textbf{\textit{f}}_j$ while fixing other vectors:
\begin{equation}
    \begin{aligned}
    &\mathop{\min}_{\textbf{\textit{f}}_j} t_{jj}\textbf{\textit{f}}_j^\top \textbf{\textit{f}}_j+(\sum\nolimits_{i=1,i\neq j}^n(t_{ij}+t_{ji})\textbf{\textit{f}}_i^\top)\textbf{\textit{f}}_j\\
    &\begin{array}{r@{\quad}r@{}l@{\quad}l}
        \mathrm{s.t.}& \textbf{\textit{f}}_j\textbf{1}_q=1,\textbf{0}_{q}\leq\textbf{\textit{f}}_j\leq\textbf{\textit{y}}_j.
    \end{array}  
    \label{eq9}
\end{aligned}
\end{equation}
Eq. \eqref{eq9} is one of the subproblems for updating $\textbf{F}$ with $q$ variables and $q+1$ constraints and it can be efficiently solved.

\begin{table*}[t]
\centering
\renewcommand{\arraystretch}{0.95}
\setlength{\tabcolsep}{1.2pt}
\resizebox{1\textwidth}{!}{\begin{tabular}{ccccc:cccc}
\Xhline{.5px}
\Xhline{.5px} 
\multicolumn{1}{c}{\multirow{2}{2cm}{\centering Compared \\ Method}} & \multicolumn{4}{c:}{\textbf{Lost}}     & \multicolumn{4}{c}{\textbf{MSRCv2}}            \\ \cline{2-9}  
& $\rho=0.01$ & $\rho=0.02$ & $\rho=0.05$ & $\rho=0.10$  & $\rho=0.01$ & $\rho=0.02$ & $\rho=0.05$ & $\rho=0.10$ \\ \cline{1-9}
\multicolumn{1}{c}{\textbf{PLC (Ours)}}      & $\mathbf{0.341\pm 0.009}$    & $\mathbf{0.353\pm0.026}$        & $\mathbf{0.399\pm0.027}$       & $\mathbf{0.497\pm0.024}$   & $\mathbf{0.329\pm 0.008}$                & $\mathbf{0.325\pm0.009}$                   & $\mathbf{0.337\pm0.026}$                       & \multicolumn{1}{c}{$\underline{0.359\pm0.020}$}        \\
\multicolumn{1}{c}{\textbf{DPCLS}}      & $\underline{0.190\pm0.029}$    & $0.219\pm0.023$        & $\underline{0.377\pm0.033}$       & $\underline{0.456\pm0.059}$   & $0.185\pm0.035$                & $0.234\pm0.038$                   & $0.299\pm0.039$                       & $0.356\pm0.029$        \\
\multicolumn{1}{c}{\textbf{AGGD}}      & $0.184\pm0.026$    & $0.216\pm0.012$        & $0.375\pm0.035$       & $0.447\pm0.040$   & $0.186\pm 0.037$                & $0.229\pm0.040$                   & $0.280\pm0.044$                       & $0.334\pm0.019$        \\
\multicolumn{1}{c}{\textbf{IPAL}}      & $0.163\pm0.019$    & $\underline{0.230\pm0.023}$        & $0.314\pm0.025$       & $0.409\pm0.046$   & $0.181\pm0.058$                & $0.227\pm0.051$                   & $0.280\pm0.025$                       & $0.346\pm0.034$        \\
\multicolumn{1}{c}{\textbf{PL-KNN}}      & $0.147\pm0.047$    & $0.170\pm0.017$        & $0.185\pm0.015$       & $0.207\pm0.015$   & $0.144\pm0.012$                & $0.171\pm0.034$                   & $0.229\pm0.033$                       & $0.273\pm0.027$        \\
\multicolumn{1}{c}{\textbf{PL-SVM}}      & $0.189\pm0.020$    & $0.192\pm0.020$        & $0.242\pm0.038$       & $0.277\pm0.029$   & $0.185\pm0.048$                & $0.202\pm0.042$                   & $0.227\pm0.046$                       & $0.287\pm0.020$        \\
\multicolumn{1}{c}{\textbf{PARM}}      & $0.112\pm0.037$    & $0.143\pm0.030$        & $0.267\pm0.024$       & $0.348\pm0.043$   & $0.225\pm0.065$                & $0.243\pm0.047$                   & $0.286\pm0.035$                       & $0.334\pm0.041$  \\
\multicolumn{1}{c}{\textbf{SSPL}}      & $0.132\pm0.063$    & $0.180\pm0.038$        & $0.253\pm0.043$       & $0.300\pm0.041$   & $\underline{0.265\pm0.061}$                & $\underline{0.267\pm0.056}$                   & $\underline{0.314\pm0.052}$                       & $\mathbf{0.386\pm0.043}$        \\ \Xhline{.5px}
\Xhline{.5px} 
\multicolumn{1}{l}{\multirow{2}{2cm}{\centering Compared \\ Method}} & \multicolumn{4}{c:}{\textbf{Mirflickr}}    &\multicolumn{4}{c}{\textbf{BirdSong}}     \\ \cline{2-9} 
& $\rho=0.01$ & $\rho=0.02$ & $\rho=0.05$ & $\rho=0.10$  & $\rho=0.01$ & $\rho=0.02$ & $\rho=0.05$ & $\rho=0.10$ \\ \cline{1-9}
\textbf{PLC (Ours)}      & $\mathbf{0.485\pm 0.041}$      &$\mathbf{0.502\pm0.040}$        & $\mathbf{0.567\pm0.021}$       &  $\mathbf{0.585\pm0.014}$   & $\mathbf{0.520\pm 0.047}$      &$\mathbf{0.539\pm0.034}$        & $\underline{0.612\pm0.018}$       &  $0.627\pm0.020$             \\
\textbf{DPCLS}    & $0.369\pm0.066$       & $\underline{0.440\pm0.046}$       & $\underline{0.507\pm0.035}$       & $\underline{0.579\pm0.034}$  & $\underline{0.492\pm0.045}$       & $0.526\pm0.035$       & $\mathbf{0.618\pm0.019}$       & $\mathbf{0.653\pm0.018}$         \\
\textbf{AGGD}     & $0.366\pm0.082$       & $0.417\pm0.055$       & $0.490\pm0.042$       & $0.555\pm0.017$  & $0.488\pm0.044$       & $\underline{0.534\pm0.037}$       & $\underline{0.612\pm0.019}$       & $\underline{0.638\pm0.014}$       \\
\textbf{IPAL}     & $0.268\pm0.098$       & $0.330\pm0.056$       & $0.426\pm0.041$       & $0.475\pm0.016$   & $0.476\pm0.043$       & $0.510\pm0.044$       & $0.586\pm0.024$       & $0.634\pm0.018$       \\
\textbf{PL-KNN}   & $0.222\pm0.056$       & $0.252\pm0.058$       & $0.324\pm0.033$       & $0.380\pm0.024$   & $0.365\pm0.072$       & $0.434\pm0.038$       & $0.532\pm0.021$       & $0.571\pm0.014$          \\
\textbf{PL-SVM}   & $0.286\pm0.054$       & $0.302\pm0.057$       & $0.436\pm0.070$       & $0.494\pm0.035$   & $0.424\pm0.054$       & $0.449\pm0.025$       & $0.559\pm0.029$       & $0.586\pm0.022$          \\ 
\textbf{PARM}     & $\underline{0.399\pm0.073}$       & $0.424\pm0.073$       & $0.451\pm0.050$       & $0.500\pm0.026$     & $0.379\pm0.029$       & $0.357\pm0.024$       & $0.353\pm0.013$       & $0.379\pm0.026$          \\ 
\textbf{SSPL}     & $0.379\pm0.084$       & $0.397\pm0.049$       & $0.399\pm0.038$       & $0.428\pm0.024$   & $0.467\pm0.036$       & $0.479\pm0.037$       & $0.521\pm0.034$       & $0.576\pm0.025$            \\ \Xhline{.5px}
\Xhline{.5px} 
\end{tabular}}
\caption{Experimental results on ACC when compared with PLL and semi-supervised PLL methods under different proportions of partial label training examples on real-world datasets, where bold and underlined indicate the best and second best results respectively.}
\label{tb4}
\end{table*}

\noindent\textbf{Update $\textbf{S}$ and $\textbf{D}$} \enspace With fixed $\textbf{W}$ and $\textbf{F}$, variable $\textbf{S}$ and $\textbf{D}$ can be updated simultaneously by solving Eq. \eqref{eq5}. To handle the equality constraint efficiently, Eq. \eqref{eq5} is approximated as:
\begin{equation}
    \begin{aligned}
    &\mathop{\min}_{\textbf{D},\textbf{S}}\parallel\textbf{D}\odot\textbf{S}\parallel_1 + \alpha \mathrm{Tr}(\textbf{D}\textbf{L}\textbf{D}^\top)+\beta \mathrm{Tr}(\textbf{S}\textbf{L}\textbf{S}^\top)\\
    &+\gamma(\parallel\textbf{P}\odot(\textbf{S}-\textbf{M})\parallel^2_F+\parallel\textbf{P}\odot(\textbf{D}-\textbf{C})\parallel^2_F)\\
    &\begin{array}{r@{\quad}l@{}l@{\quad}l}
         \mathrm{s.t.} &\textbf{S}\geq\textbf{0}_{n\times n}, \textbf{D}\geq\textbf{0}_{n\times n},\\
    \end{array}  
    \label{eq10}
\end{aligned}
\end{equation}
where $\textbf{P}$ represents the position of PCs, and $\textbf{P}_{ij}=1$ if $(\textbf{\textit{x}}_i, \textbf{\textit{x}}_j)\in \mathcal{M}\cup\mathcal{C}$, otherwise $\textbf{P}_{ij}=0$. Eq. \eqref{eq10} is convex to $\textbf{S}$ (or $\textbf{D}$) when $\textbf{D}$ (or $\textbf{S}$) is fixed. We can introduce the Lagrange multiplier matrices $\Phi_{\textbf{S}}\in\mathbb{R}^{n\times n}$ and $\Phi_{\textbf{D}}\in\mathbb{R}^{n\times n}$ to deal with the non-negative constraint and the Lagrangian function is expressed as:
\begin{equation}
    \begin{aligned}
    \mathcal{L}=&\parallel\textbf{D}\odot\textbf{S}\parallel_1 + \alpha \mathrm{Tr}(\textbf{D}\textbf{L}\textbf{D}^\top)+\beta \mathrm{Tr}(\textbf{S}\textbf{L}\textbf{S}^\top)\\
    &+\gamma(\parallel\textbf{P}\odot(\textbf{S}-\textbf{M})\parallel^2_F+\parallel\textbf{P}\odot(\textbf{D}-\textbf{C})\parallel^2_F)\\
    &+\mathrm{Tr}(\Phi_{\textbf{S}}^\top\textbf{S})+\mathrm{Tr}(\Phi_{\textbf{D}}^\top\textbf{D}),
    \label{eq11}
\end{aligned}
\end{equation}
according to the Karush–Kuhn–Tucker (KKT) conditions, we have $\textbf{S}\odot\Phi_{\textbf{S}}=\textbf{0}$ and $\textbf{D}\odot\Phi_{\textbf{D}}=\textbf{0}$. Let $\partial\mathcal{L}/\partial\textbf{S}_{ij}=0$ and $\partial\mathcal{L}/\partial\textbf{D}_{ij}=0$, the updating formula are shown as follows:
\begin{equation}
    \begin{aligned}
    \textbf{S}_{ij}=\textbf{S}_{ij}\frac{(2\gamma\textbf{P}\odot\textbf{M}+2\beta\textbf{S}\textbf{W})_{ij}}{(\textbf{D}+2\beta\textbf{S}\textbf{A}+2\gamma\textbf{P}\odot\textbf{S})_{ij}},
    \label{eq12}
\end{aligned}
\end{equation}

\begin{equation}
    \begin{aligned}
    \textbf{D}_{ij}=\textbf{D}_{ij}\frac{(2\gamma\textbf{P}\odot\textbf{C}+2\alpha\textbf{D}\textbf{W})_{ij}}{(\textbf{S}+2\alpha\textbf{D}\textbf{A}+2\gamma\textbf{P}\odot\textbf{D})_{ij}}.
    \label{eq13}
\end{aligned}
\end{equation}
After alternating optimization, we obtain a weight matrix that can fully represent the similarity relationship of the samples. Finally, we apply SC on the weight matrix $\textbf{W}$ to get the final clustering results.

The overall pseudo-code of our PLC method is summarized in \textbf{Algorithm \ref{alg:algorithm}} and the computational complexity analysis of \textbf{Algorithm \ref{alg:algorithm}} can be found in \textbf{Appendix C}.


\section{Theoretical Analysis} 
\begin{theorem}
    Denote $\textbf{F}\in[0, 1]^{n\times q}$ and $\textbf{W}\in[0, 1]^{n\times n}$ as the label confidence matrix and the weight matrix to be optimized. Let $\textbf{F}_G$ and $\textbf{W}_G$ be the ground-truth label matrix and the optimal weight matrix under the ground-truth labels. We assume that $\textbf{W}_G$ is constructed on the premise that the ground-truth labels of neighboring examples are the same. Let $\lambda$ be the smallest eigenvalue of $\textbf{F}^\top_G\textbf{F}_G$ and $||\overline{\Delta}_{\textbf{W}}||_F$ be the average distance of each corresponding position between $\textbf{W}_G$ and $\textbf{W}$, i.e., $||\overline{\Delta}_{\textbf{W}}||_F=\frac{1}{n^2}||\textbf{W}_G-\textbf{W}||_F$. Then we have
    \begin{equation}
        \begin{aligned}
            ||\overline{\Delta}_{\textbf{W}}||_F\leq\frac{n+2}{\lambda n}||\textbf{F}^\top\textbf{F}-\textbf{F}^\top_G\textbf{F}_G||_F+\frac{2n+q-1}{\lambda n}.
        \label{eq14}
        \end{aligned}
    \end{equation}
    \label{theorem1}
\end{theorem}

The proof can be found in \textbf{Appendix B}. From {\bf Theorem 1}, we find that a smaller difference between $\textbf{F}$ and $\textbf{F}_G$ can reduce the upper bound of $||\overline{\Delta}_{\textbf{W}}||_F$, indicating that better disambiguation results help achieve a better weight matrix, thereby improving clustering performance.{\it In summary, we prove that, under some general assumptions, a better disambiguated label matrix improves clustering performance.}


\section{Experiments}
\subsection{Experimental Setup}
\textbf{Datasets} \enspace To conduct a comprehensive evaluation of our proposed method, we compare our PLC method with other methods on both controlled UCI datasets and real-world datasets. The characteristics of controlled UCI datasets and real-world datasets can be found in \textbf{Appendix D}. Following the widely-used partial label data generation protocol \cite{2011Learning}, we generate the artificial partial label datasets under the controlling parameter $r$ which controls the number of false-positive labels. For each example, we randomly select $r$ other labels as false-positive labels. 



\begin{table*}[t]
\small
\renewcommand{\arraystretch}{0.9}
\centering
\resizebox{1\textwidth}{!}{\begin{tabular}{ccccccc}
\Xhline{.5px}
\Xhline{.5px} 
\multicolumn{1}{l}{\multirow{2}{2cm}{\centering Compared \\ Method}} & \multicolumn{6}{c}{\textbf{Lost}}                   \\ \cline{2-7} 
\multicolumn{1}{l}{}  & $\rho=0.05$ & $\rho=0.10$ & $\rho=0.15$ & $\rho=0.20$ & $\rho=0.30$ & $\rho=0.40$  \\ \hline
\textbf{PLC}      & $\mathbf{0.399\pm0.027}$       & $\mathbf{0.497\pm0.024}$  & $\mathbf{0.512\pm0.024}$       & $\mathbf{0.553\pm0.026}$    & $\mathbf{0.608\pm0.021}$       & $\mathbf{0.641\pm0.022}$      \\
\textbf{PLC-CW}     &$0.349\pm0.021$        &$0.352\pm0.015$        &$0.349\pm0.021$        &$0.351\pm0.021$   &$0.361\pm0.018$  &$0.348\pm0.015$              \\
\textbf{PLC-LD}     &$0.355\pm0.021$        &$0.379\pm0.019$       &$0.371\pm0.024$       &$0.393\pm0.018$   &$0.395\pm0.016$  &$0.402\pm0.014$             \\
\textbf{PLC-SD}    & $0.350\pm0.021$       & $0.344\pm0.032$       & $0.307\pm0.026$       & $0.265\pm0.026$   &$0.211\pm0.011$ &$0.211\pm0.016$           \\ \Xhline{.5px}
\Xhline{.5px} 
\multicolumn{1}{l}{\multirow{2}{2cm}{\centering Compared \\ Method}} & \multicolumn{6}{c}{\textbf{MSRCv2}}                   \\ \cline{2-7} 
\multicolumn{1}{l}{}  & $\rho=0.05$ & $\rho=0.10$ & $\rho=0.15$ & $\rho=0.20$ & $\rho=0.30$ & $\rho=0.40$  \\ \hline
\textbf{PLC}      & $\mathbf{0.337\pm 0.026}$      &$\mathbf{0.359\pm0.020}$        & $\mathbf{0.372\pm0.021}$       &  $\mathbf{0.404\pm0.016}$  &$\mathbf{0.425\pm0.015}$ &$\mathbf{0.433\pm0.017}$     \\
\textbf{PLC-CW}     &$0.318\pm0.015$        &$0.315\pm0.007$        &$0.317\pm0.014$        &$0.320\pm0.011$   &$0.318\pm0.012$  &$0.317\pm0.013$              \\
\textbf{PLC-LD}     &$0.322\pm0.015$        &$0.320\pm0.014$       &$0.321\pm0.009$       &$0.323\pm0.013$   &$0.324\pm0.014$  &$0.342\pm0.016$             \\
\textbf{PLC-SD}    & $0.319\pm0.009$       & $0.313\pm0.013$       & $0.297\pm0.017$       & $0.270\pm0.021$   &$0.209\pm0.028$ &$0.198\pm0.015$           \\ \Xhline{.5px}
\Xhline{.5px} 
\end{tabular}}
\caption{Ablation study of the proposed method PLC, where bold indicates the best result.}
\label{tb5}
\end{table*}

\begin{figure*}[t]
        \centering
        \begin{minipage}{0.27\linewidth}
		  \vspace{3pt}
            \centerline{\includegraphics[width=\linewidth]{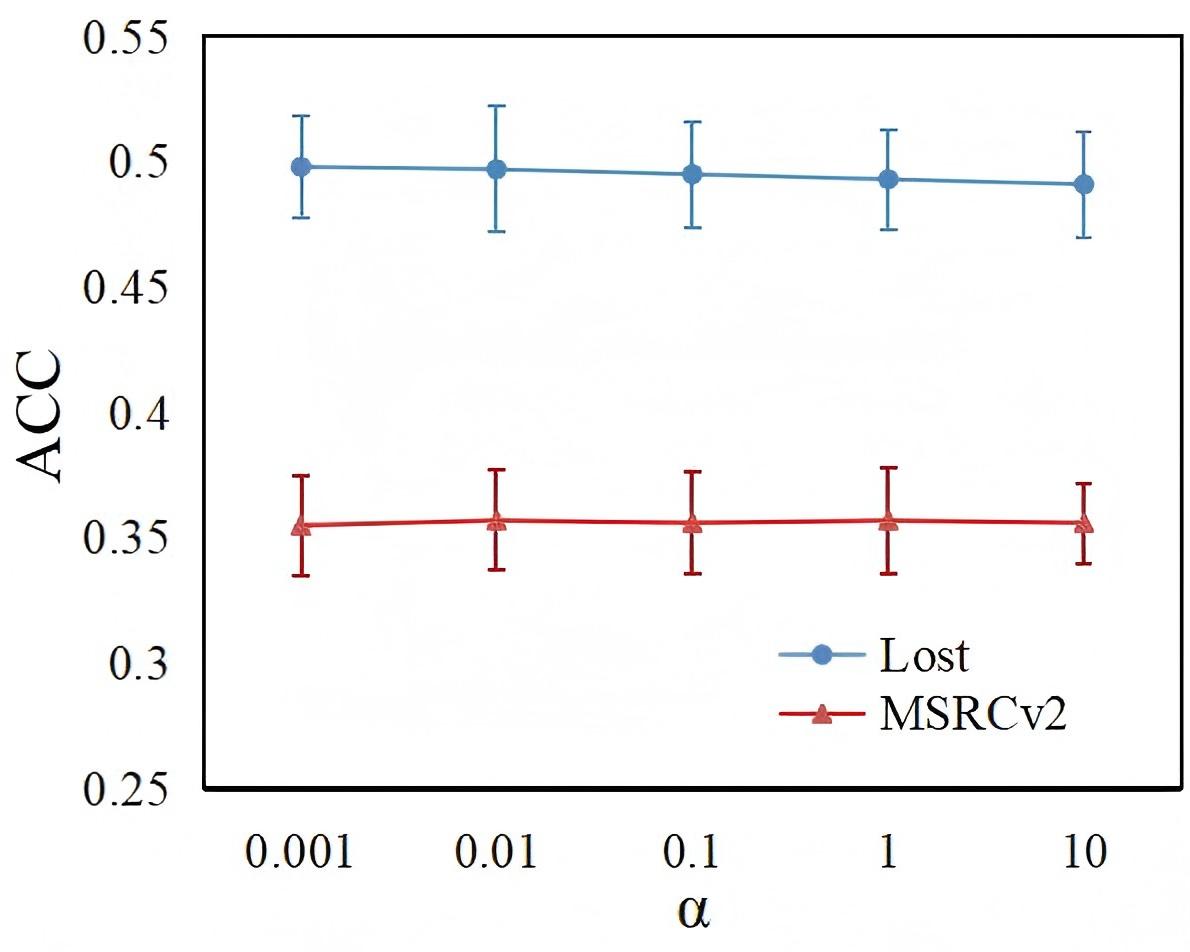}}
            \centerline{ \enspace  (a) Varying $\alpha$}
        \end{minipage}
        \hspace{1em}
        \begin{minipage}{0.27\linewidth}
            \vspace{3pt}
            \centerline{\includegraphics[width=\linewidth]{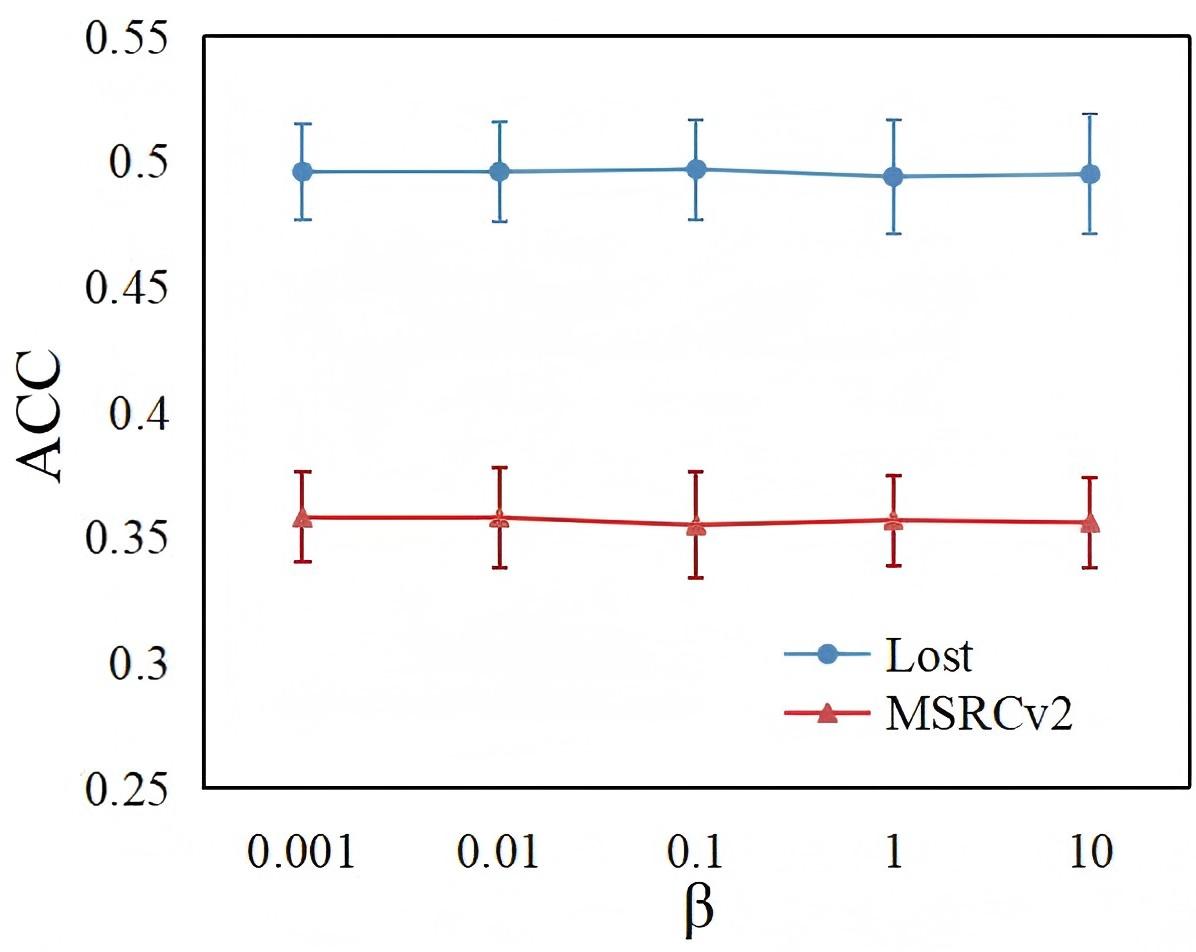}}
            \centerline{ \enspace  (b) Varying $\beta$}
        \end{minipage}
        \hspace{1em}
        \begin{minipage}{0.27\linewidth} 
            \vspace{3pt}
            \centerline{\includegraphics[width=\linewidth]{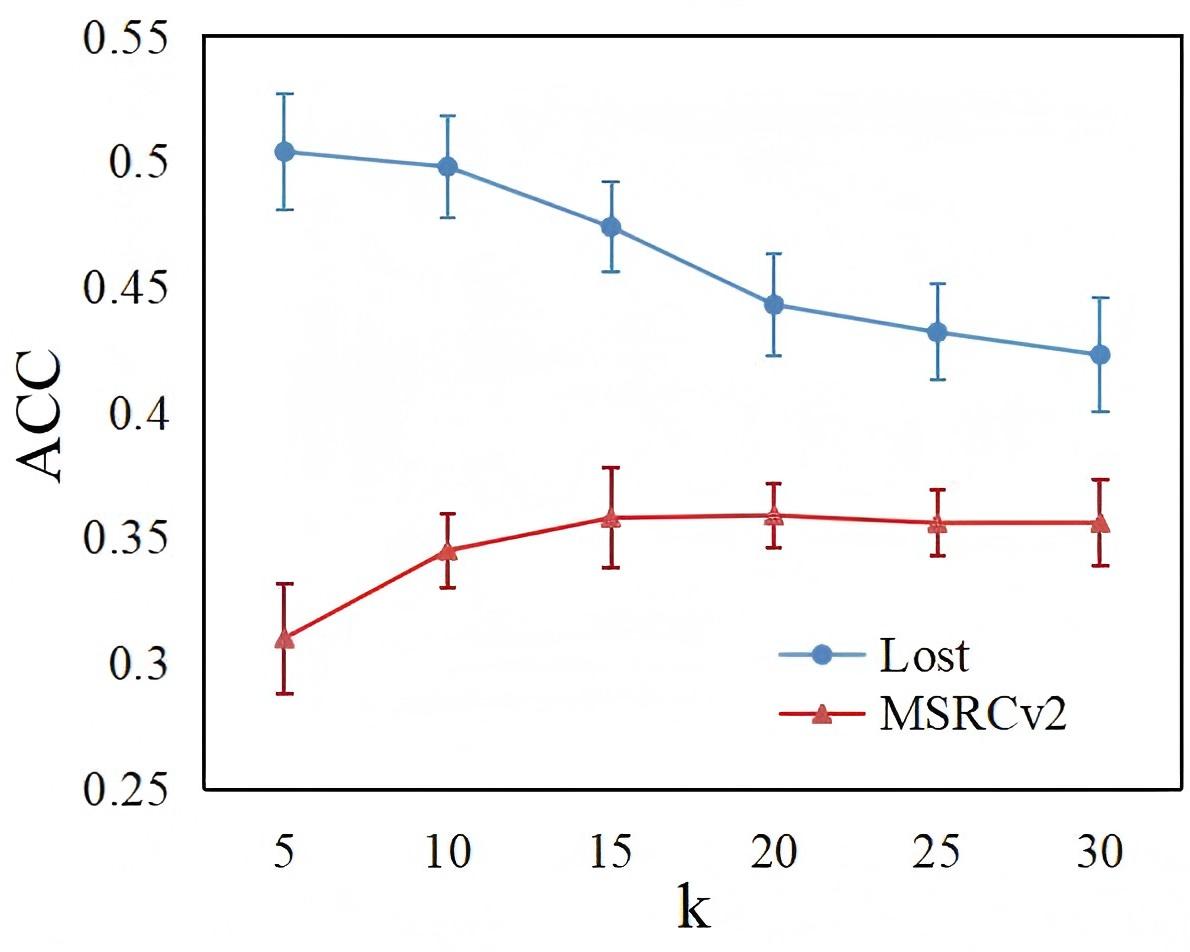}}
            \centerline{ \enspace  (c) Varying $k$}
        \end{minipage}
    \caption{Parameter sensitivity analysis for PLC. (a) ACCs of PLC on Lost and MSRCv2 by varying $\alpha$; (b) ACCs of PLC on Lost and MSRCv2 by varying $\beta$; (c) ACCs of PLC on Lost and MSRCv2 by varying $k$;}
    \label{f3}
\end{figure*}

\noindent\textbf{Compared Methods} \enspace To demonstrate the effectiveness of our PLC method, we compare it with two baseline clustering methods, three state-of-the-art constrained clustering methods, five well-established PLL methods, and two semi-supervised PLL methods. (1) Baseline clustering method: K-Means \cite{1967Some} and SC \cite{Ng2001OnSC}. (2) Constrained clustering methods: SSC-TLRR \cite{10007868}, DP-GLPCA \cite{9178787} and SSSC \cite{2018Semi}. (3) PLL methods: PL-KNN \cite{2009Learning}, PL-SVM \cite{Nguyen2008ClassificationWP}, DPCLS \cite{NEURIPS2023_6b97236d}, AGGD \cite{9573413} and IPAL \cite{Zhang2015SolvingTP}. (4) Semi-supervised PLL methods: SSPL \cite{Wang2019PartialLL} and PARM \cite{Wang2020SemiSupervisedPL}. Each compared method is implemented with the default hyper-parameter setup suggested in the respective literature. Parameters for our PLC method are set as $\alpha,\beta\in\{0.01,0.1,1\}$, $\gamma=10$ and $k\in\{10,15,20,25,30,40\}$.

\noindent\textbf{Implementation Details} \enspace For constrained clustering methods, we randomly sample the partial label examples based on the proportion $\rho \in\{0.05,0.10,0.15,0.20,0.30,0.40\}$ and the remaining samples are used as test data. For PLL and semi-supervised PLL methods, we randomly sample the partial label examples based on the proportion $\rho \in\{0.01,0.02,0.05,0.10\}$. For each experiment, we implemented 10 times with random partitions and reported the average performance with the standard deviation. We used ACC and NMI as evaluation metrics. For detailed information, please refer to \textbf{Appendix A}.



\subsection{Experimental Results}
\subsubsection{Comparison with Constrained Clustering Methods}
Fig. \ref{f1} illustrates the ACCs and the NMIs of our PLC method compared to constrained clustering methods under different proportions of partial label training examples on synthetic UCI datasets. According to these figures, our PLC method ranks first in 88.9\% (64/72) cases. Table \ref{tb2} reports the ACCs of our PLC method compared to constrained clustering methods under different proportions of partial label training examples on real-world datasets. Our PLC method ranks first in 100\% (24/24) cases. Moreover, we can find that constrained clustering methods perform poorly in some cases, and there is a decline in performance despite an increase in the proportion of partial labeled samples, indicating that constrained clustering methods are difficult to handle the ambiguous labels. In contrast, our PLC method performs better in most cases, indicating that the PLC method can effectively utilize the partial label samples to predict the unlabeled samples. 
\subsubsection{Comparison with PLL \& Semi-supervised PLL Methods}
Table \ref{tb4} reports the ACCs of our PLC method compared to PLL and semi-supervised PLL methods under different proportions of partial label training examples on real-world datasets. Our PLC method achieves superior performance against PL-KNN, PL-SVM, IPAL, PARM in 100\% (16/16) cases, against SSPL in 93.75\% (15/16) cases, and against AGGD and DPCLS in 87.5\% (14/16) cases. We can find that our PLC method performs better in the case of fewer partial labeled samples than PLL methods and semi-supervised PLL methods, indicating that our PLC method is helpful in addressing the issue of insufficient partial labels.

More experimental results can be found in \textbf{Appendix E}. We also conduct a significance analysis to prove that our PLC method is significantly superior to the comparing methods. The results can be found in \textbf{Appendix E.4}.


\subsection{Further Analysis} 
\textbf{Ablation Study.} \enspace In Table \ref{tb5}, we conduct an ablation study on Lost and MSRCv2 datasets with different proportion $\rho\in\{0.05,0.10,0.15,0.20,0.30,0.40\}$ to check the necessity of the terms involved in our method. Specifically, \textbf{PLC-CW} represents the \textbf{PLC} version that only utilizes sample features to construct a weight matrix and perform the SC, \textbf{PLC-LD} represents the \textbf{PLC} version that utilizes sample features and disambiguated partial labels to construct a weight matrix and perform SC, and \textbf{PLC-SD} represents the \textbf{PLC} version that utilizes pairwise constraint propagation to densify the initial pairwise constraint and perform SC. The results in Table \ref{tb5} indicate that label disambiguation and pairwise constraint propagation are helpful in improving classification accuracy, and taking both into account is the best choice.

\noindent\textbf{Parameter Sensitivity} \enspace Fig. \ref{f3} reports the performance of our PLC method under different parameter configurations on datasets Lost and MSRCv2. According to Fig. \ref{f3}, the performance of our PLC method is relative stable in the different settings of $\alpha$ and $\beta$ (Fig. \ref{f3}(a) and Fig. \ref{f3}(b)). We can directly fix $\alpha,\beta=0.1$ in practice. In addition, the accuracy curves in Fig. \ref{f3}(c) indicate that our PLC is relative sensitive to the parameter $k$. In practice, we find that setting a smaller $k\in\{10,15\}$ for small-scale datasets and a larger $k\in\{20,30,40\}$ for large-scale datasets can improve the performance of our PLC method.

\section{Conclusion}
In this paper, we proposed a novel method named PLC. For the first time, we explore whether partial labels can help improve the performance of clustering methods. Specifically, we first construct a weight matrix based on the relationships in the feature space and disambiguate the candidate labels to estimate the ground-truth label. Then we establish the must-link constraint and the cannot-link constraint based on the disambiguation results. Finally, we propagate the initial pairwise constraints and form an adversarial relationship between these two constraints to improve clustering performance. Moreover, we theoretically prove that a better disambiguated label matrix improves clustering performance, confirming that using label disambiguation to enhance clustering performance is effective. Our PLC method achieves superior performance compared to conventional constrained clustering, PLL and semi-supervised PLL methods on both three synthetic datasets and four real-world datasets.


\section*{Acknowledgments}
This research work is supported by the Big Data Computing Center of Southeast University.

\bibliographystyle{named}
\bibliography{ijcai25}

\clearpage

\appendix

\section{Evaluation Metrics}
We use Average Clustering Accuracy (ACC) and Normalized Mutual Information (NMI) metrics, both of which are widely used criteria in the field of clustering. ACC discovers the one-to-one relationship between clusters and classes. Denote $c_i$ as the clustering result of sample $\textbf{\textit{x}}_i$ and $g_i$ as the ground-truth label of sample $\textbf{\textit{x}}_i$, ACC is defined as
 \begin{equation}
     \begin{aligned}
     ACC=\frac{1}{n}\sum_{i=1}^n\delta(c_i,map(g_i)), 
     \label{eq.S1}
 \end{aligned}
 \end{equation}
 where $\delta(p,q)=1$, if $p=q$ and $\delta(p,q)=0$ otherwise. $n$ is the total number of examples and $map(g_i)$ is the mapping function that permutes the clusters to match the ground-truth labels. NMI measures the mutual information entropy between the clusters and the ground-truth labels. Given the ground-truth labels $\textbf{Y}$ and the clustering results $\textbf{C}$, NMI is defined as
 \begin{equation}
     \begin{aligned}
     NMI=\frac{\sum_{y\in\textbf{Y}, c\in\textbf{C}}p(y,c)log(\frac{p(y,c)}{p(y)p(c)})}{\sqrt{\sum_{y\in\textbf{Y}}p(y)\log p(y)\sum_{c\in\textbf{C}}p(c)\log p(c)}}, 
     \label{eq.S2}
 \end{aligned}
 \end{equation}
 where $p(y)$ and $p(c)$ represent the marginal probability distribution functions of $\textbf{Y}$ and $\textbf{C}$ respectively, and $p(y,c)$ is the joint distribution.

\section{Proof of Theorem 1}
We first give a lemma as follows.
\newtheorem{lemma}{Lemma}
\begin{lemma}
\begin{equation}
    \begin{aligned}
       \mathrm{Tr}(\textbf{A}\textbf{B})\leq||\textbf{A}||_F||\textbf{B}||_F.
    \end{aligned}
    \label{eq.S10}
\end{equation}
\begin{proof}
    By Cauchy-Schwarz, we have
    \begin{equation}
    \begin{aligned}
       \mathrm{Tr}(\textbf{A}\textbf{B})&=\sum_{i,j}a_{ij}b_{ji}\\
       &\leq\sum_i(\sum_j|a_{ij}|^2)^{1/2}(\sum_j|b_{ji}|^2)^{1/2}\\
       &\leq(\sum_{i,j}|a_{ij}|^2)^{1/2}(\sum_{i,j}|b_{ji}|^2)^{1/2}\\
       &=||\textbf{A}||_F||\textbf{B}||_F.
    \end{aligned}
    \label{eq.S11}
\end{equation}
This concludes the proof.
\end{proof}
\end{lemma}
\noindent Now we begin the proof of Theorem 1.
\begin{proof}
Denote $\textbf{F}\in[0, 1]^{n\times q}$ and $\textbf{W}\in[0, 1]^{n\times n}$ the label confidence matrix and the weight matrix to be optimized. For the convenience of explanation, the terms related to $\textbf{F}$ in the objective function Eq. (2) can be rewritten as
\begin{equation}
    \begin{aligned}
        &\min_{\textbf{W}}||\textbf{F}-\textbf{W}\textbf{F}||^2_F\\
        &\begin{array}{l@{\quad}l@{}l@{\quad}l@{\quad}}
         \mathrm{s.t.} 
         &w_{ij}=0 \enspace\mathrm{if} \enspace(\textbf{\textit{x}}_i, \textbf{\textit{x}}_j)\notin\mathcal{E},\\
         &\textbf{W}^\top\textbf{1}_n=\textbf{1}_n,\textbf{0}_{n\times n}\leq\textbf{W}\leq\textbf{N}.\\
    \end{array} 
    \end{aligned}
    \label{eq.S3}
\end{equation}
Let $\textbf{F}_G$ and $\textbf{W}_G$ be the ground-truth label matrix and the optimal weight matrix under the ground-truth labels. We assume that $\textbf{W}_G$ is constructed on the premise that the ground-truth labels of neighboring examples are the same, which can improve clustering performance. Due to the constraint of $\textbf{W}_G^\top\textbf{1}_n=\textbf{1}_n$, we have $||\textbf{F}_G-\textbf{W}_G\textbf{F}_G||^2_F=0$. Denote $\Delta_{\textbf{W}}=\textbf{W}_G-\textbf{W}$, the following inequality holds
\begin{equation}
    \begin{aligned}
        &||\textbf{F}_G-(\Delta_{\textbf{W}}+\textbf{W})\textbf{F}_G||^2_F\leq||\textbf{F}-\textbf{W}\textbf{F}||^2_F.
    \end{aligned}
    \label{eq.S4}
\end{equation}

Expand Eq. \eqref{eq.S4}, we have
\begin{equation}
    \begin{aligned}
        &||\Delta_{\textbf{W}}\textbf{F}_G||^2_F\\
        &\leq||\textbf{F}||^2_F+\mathrm{Tr}(\textbf{W}^\top\textbf{W}(\textbf{F}^\top\textbf{F}-\textbf{F}_G^\top\textbf{F}_G))\\
        &\quad+\mathrm{Tr}((\textbf{W}+\textbf{W}^\top)(\textbf{F}_G^\top\textbf{F}_G-\textbf{F}^\top\textbf{F}))-||\textbf{F}_G||^2_F\\
        &\quad+\mathrm{Tr}(\textbf{F}_G^\top\textbf{F}_G((\textbf{I}-\textbf{W})^\top\Delta_{\textbf{W}}+(\textbf{I}-\textbf{W})\Delta_{\textbf{W}}^\top)).\\
    \end{aligned}
    \label{eq.S5}
\end{equation}

According to Lemma 1 and the fact that the Frobenius norm is submultiplicative, we have
\begin{equation}
    \begin{aligned}
        &||\Delta_{\textbf{W}}\textbf{F}_G||^2_F\\
        &\leq||\textbf{F}||^2_F-||\textbf{F}_G||^2_F+2||\textbf{F}_G||^2_F||\textbf{I}-\textbf{W}||_F||\Delta_{\textbf{W}}||_F\\
        &\quad(||\textbf{W}||^2_F+2||\textbf{W}||_F)||\textbf{F}^\top\textbf{F}-\textbf{F}_G^\top\textbf{F}_G||_F.\\
    \end{aligned}
    \label{eq.S12}
\end{equation}

Since $\textbf{W}$ is upper bounded by the number of samples $n$, we have $||\textbf{W}||^2_F\leq n^2$ and $||\textbf{I}-\textbf{W}||^2_F\leq n^2$. Due to $\textbf{F}_G$ is the ground-truth label matrix, we have $||\textbf{F}_G||_F^2=n$. Furthermore, $\textbf{F}$ is upper bounded by the number of samples $n$ and the number of classes $q$, i.e., $||\textbf{F}||^2_F\leq nq$. Assume that $||\Delta_{\textbf{W}}||_F\geq1$, which is a reasonable assumption when $n$ is large enough. We have

\begin{equation}
    \begin{aligned}
        ||\Delta_{\textbf{W}}\textbf{F}_G||^2_F\leq&(n^2+2n)||\textbf{F}^\top\textbf{F}-\textbf{F}_G^\top\textbf{F}_G||_F||\Delta_{\textbf{W}}||_F\\
        &+(2n^2+nq-n)||\Delta_{\textbf{W}}||_F.
    \end{aligned}
    \label{eq.S6}
\end{equation}

Note that $\textbf{F}^\top_G\textbf{F}_G$ is a positive semidefinite matrix, thus its eigenvalues are non-negative. Taking $\lambda$ as the smallest eigenvalue of $\textbf{F}^\top_G\textbf{F}_G$, we have $\lambda||\Delta_{\textbf{W}}||^2_F\leq||\Delta_{\textbf{W}}\textbf{F}_G||^2_F$. Thus, Eq. \eqref{eq.S6} can be further relaxed as
\begin{equation}
    \begin{aligned}
        \lambda||\Delta_{\textbf{W}}||^2_F\leq&(n^2+2n)||\textbf{F}^\top\textbf{F}-\textbf{F}_G^\top\textbf{F}_G||_F||\Delta_{\textbf{W}}||_F\\
        &+(2n^2+nq-n)||\Delta_{\textbf{W}}||_F.
    \end{aligned}
    \label{eq.S7}
\end{equation}

Let $||\overline{\Delta}_{\textbf{W}}||_F$ be the average distance of each corresponding position between $\textbf{W}_G$ and $\textbf{W}$, i.e., $||\overline{\Delta}_{\textbf{W}}||_F=\frac{1}{n^2}||\textbf{W}_G-\textbf{W}||_F$. Dividing $n^2$ on both sides of Eq. \eqref{eq.S7}, we finally have
\begin{equation}
        \begin{aligned}
            ||\overline{\Delta}_{\textbf{W}}||_F\leq\frac{n+2}{\lambda n}||\textbf{F}^\top\textbf{F}-\textbf{F}^\top_G\textbf{F}_G||_F+\frac{2n+q-1}{\lambda n}.
        \end{aligned}
        \label{eq.S8}
\end{equation}
This concludes the proof of Theorem 1.
\end{proof}


\begin{table*}[hp]
 
 \centering
 \resizebox{0.75\textwidth}{!}{%
 \begin{tabular}{ccccc}
 \Xhline{.5px}
 \Xhline{.5px} 
 \multicolumn{5}{l}{\textbf{Controlled UCI Datasets}}                                                                                                 \\ \hline 
 \textbf{Dataset} & \textbf{\# Examples} & \textbf{\# Features} & \textbf{\# Class Labels} & \multicolumn{1}{l}{\textbf{\# False Positive Labels ($r$)}} \\ \hline
 \textbf{Ecoli}   & 336                  & 7                    & 8                        & $r=1,2,3$                                                   \\
 \textbf{Vehicle}   & 846                 & 18                    & 4                        & $r=1,2$                                                  \\
 \textbf{Coil20}   & 1440                  & 1024                   & 20                     & $r=1,2,3$                                                \\ \Xhline{.5px}
 \Xhline{.5px} 
 \end{tabular}%
 }

 \vspace{8pt}

\resizebox{0.85\textwidth}{!}{%

 \begin{tabular}{cccccc}
 \Xhline{.5px}
 \Xhline{.5px} 
 \multicolumn{6}{l}{\textbf{Real-World Datasets}}                                                                                                                                 \\ \hline 
 \textbf{Dataset}   & \textbf{\# Examples} & \textbf{\# Features} & \textbf{\# Class Labels} & \multicolumn{1}{l}{\textbf{Avg.\# CLs}} & \multicolumn{1}{c}{\textbf{Task Domain}} \\ \hline 
 \textbf{Lost}      & 1122                 & 108                  & 16                       & 2.23                                    & automatic face naming                    \\
 \textbf{MSRCv2}    & 1758                 & 48                   & 23                       & 3.16                                    & object classification                    \\
 \textbf{Mirflickr} & 2780                 & 1536                 & 14                       & 2.76                                    & web image classification                 \\
 \textbf{BirdSong}  & 4998                 & 38                   & 13                       & 2.18                                    & bird song classification                 \\ 
 \textbf{LYN10}  & 16526                 & 163                   & 10                       & 1.84                                    & automatic face naming                 \\ 
 \Xhline{.5px}
 \Xhline{.5px} 
 \end{tabular}%
 }
 \caption{Characteristics of controlled UCI datasets and real-world datasets.}
 \label{tb1}
 \end{table*}

\begin{figure*}[hp]
    \centering
    \subfigure{
    \includegraphics[width=3.4cm,height=3.4cm]{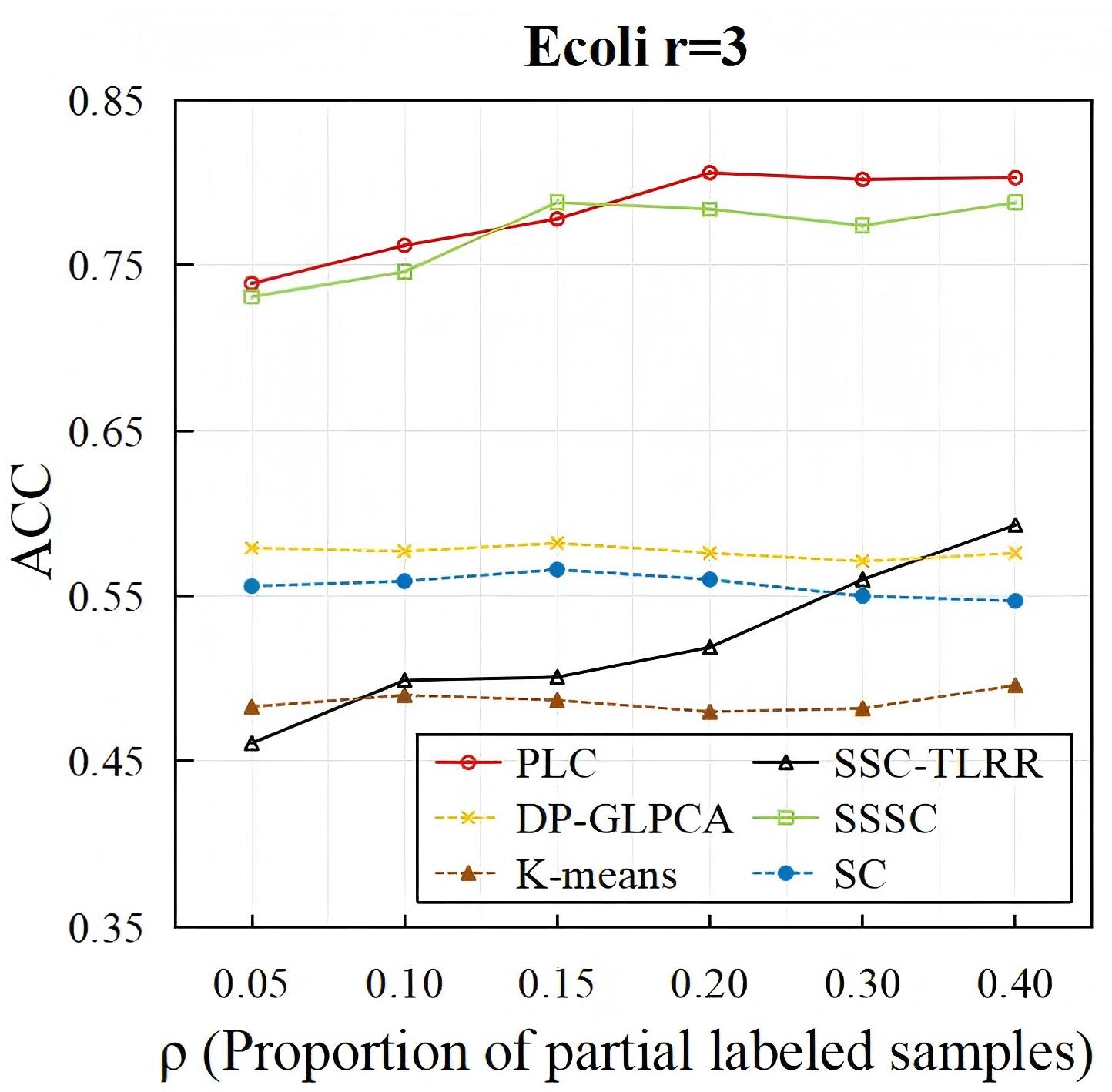} }
    \hspace{2.5mm}
    \subfigure{
    \includegraphics[width=3.4cm,height=3.4cm]{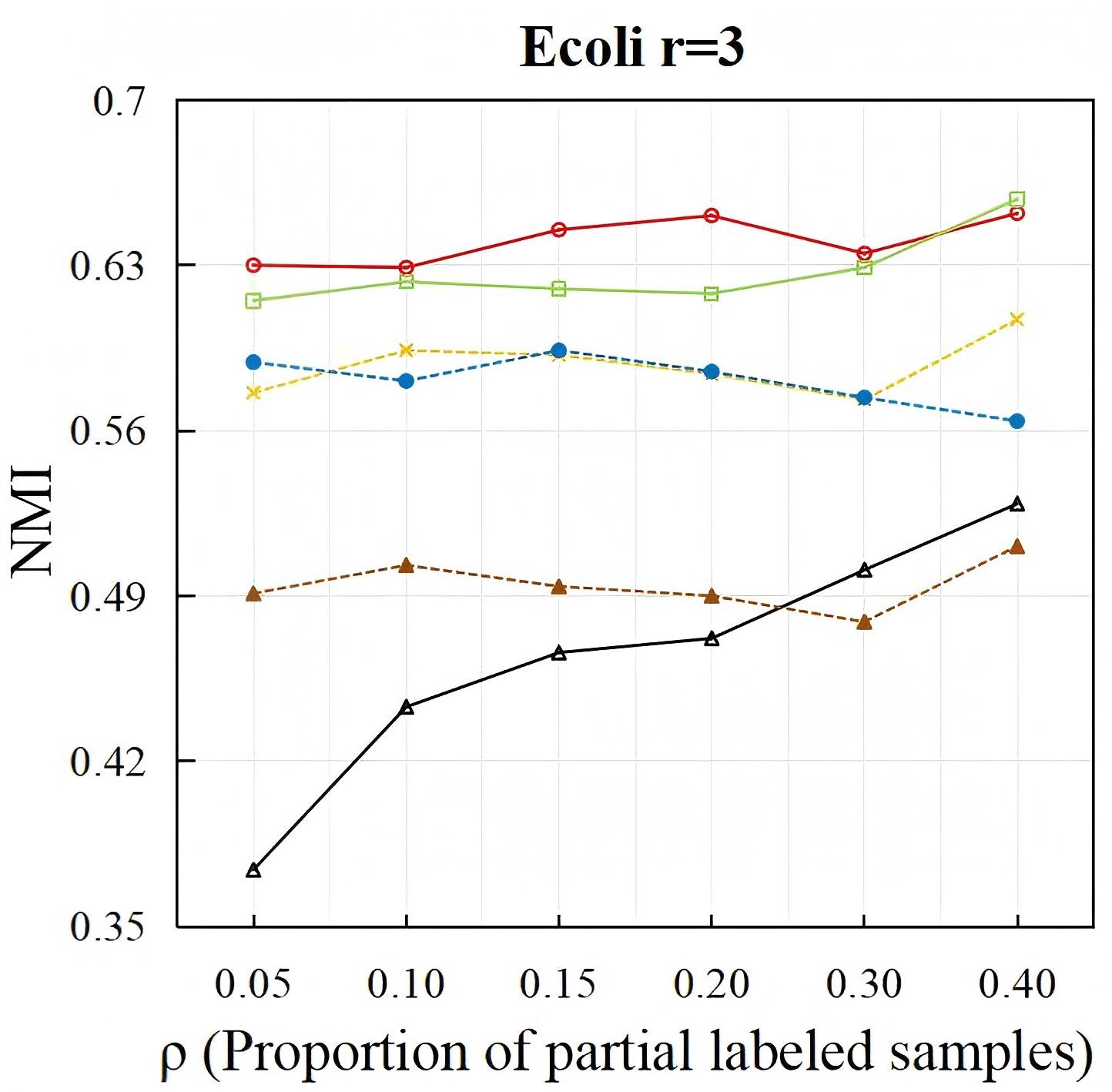} }	
    \hspace{2.5mm}
    \subfigure{
    \includegraphics[width=3.4cm,height=3.4cm]{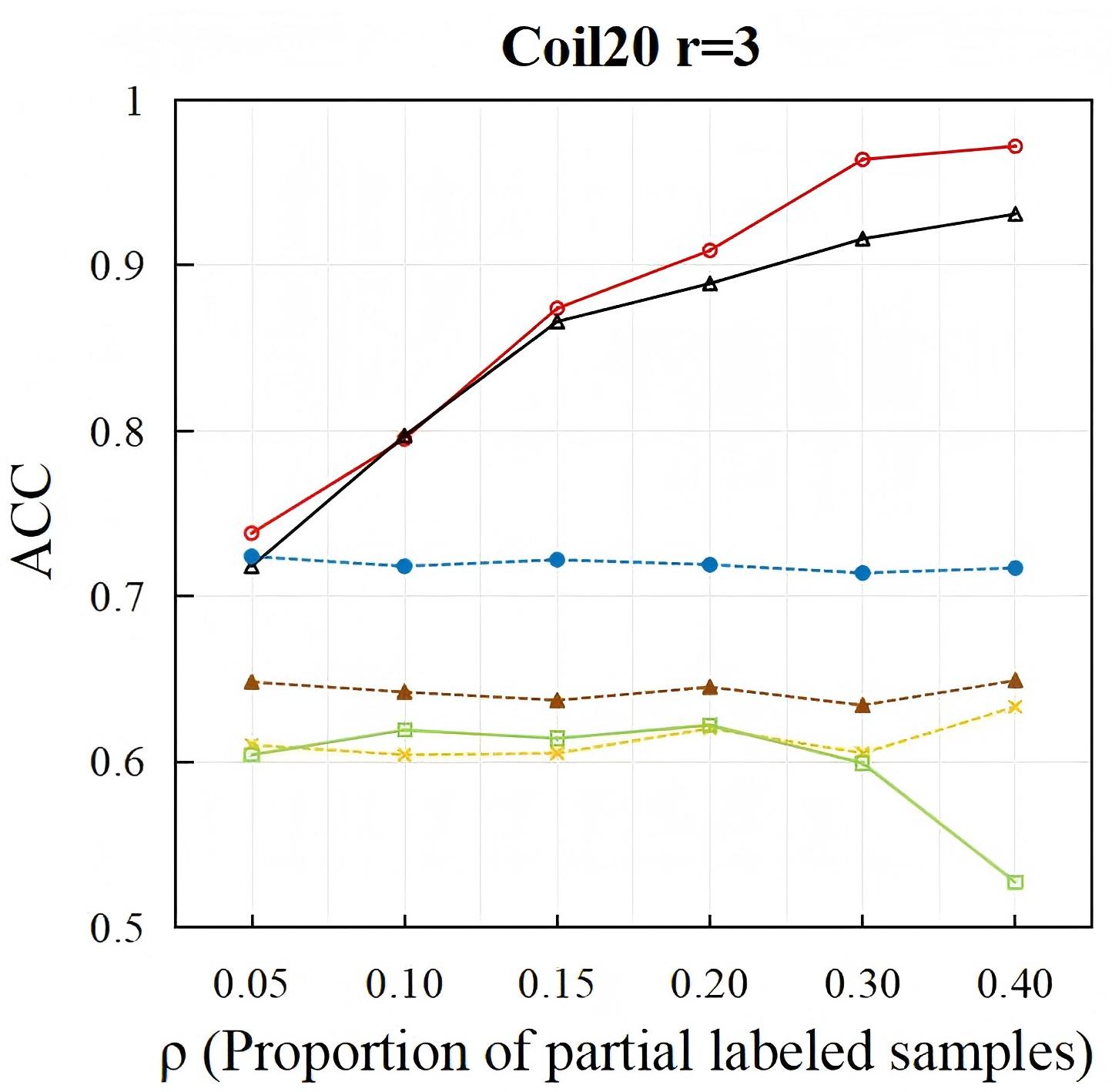} }
    \hspace{2.5mm}
    \subfigure{
    \includegraphics[width=3.4cm,height=3.4cm]{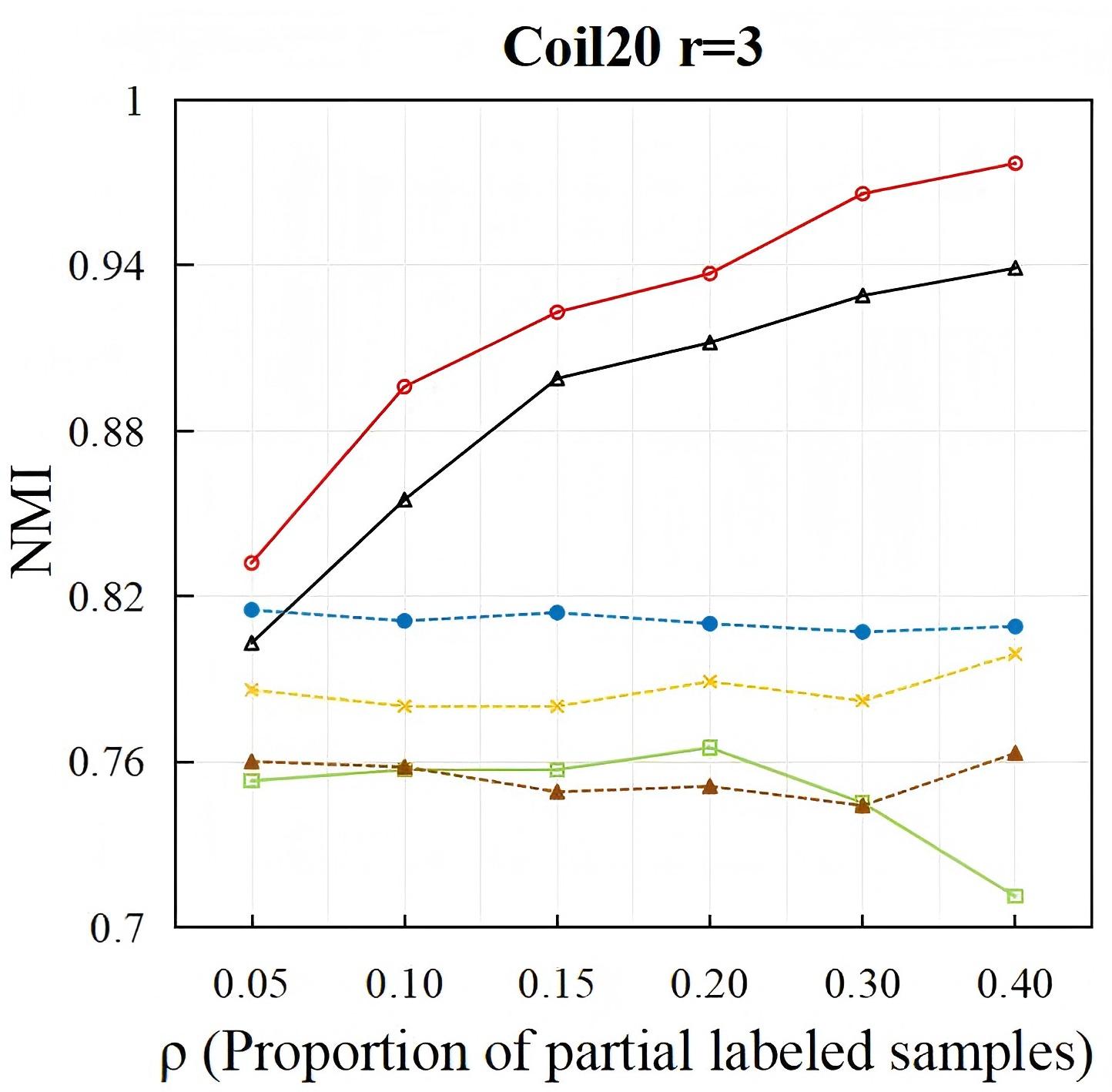} }
    \hspace{2.5mm}
    \caption{ACCs and NMIs when compared with constrained clustering methods under different proportions of partial label training samples on the datasets Ecoli $r=3$ and Coil20 $r=3$. }
    \label{f2}
\end{figure*}

\begin{figure*}[hp]
     \centering
    \subfigure{
    \includegraphics[width=3.5cm,height=3.5cm]{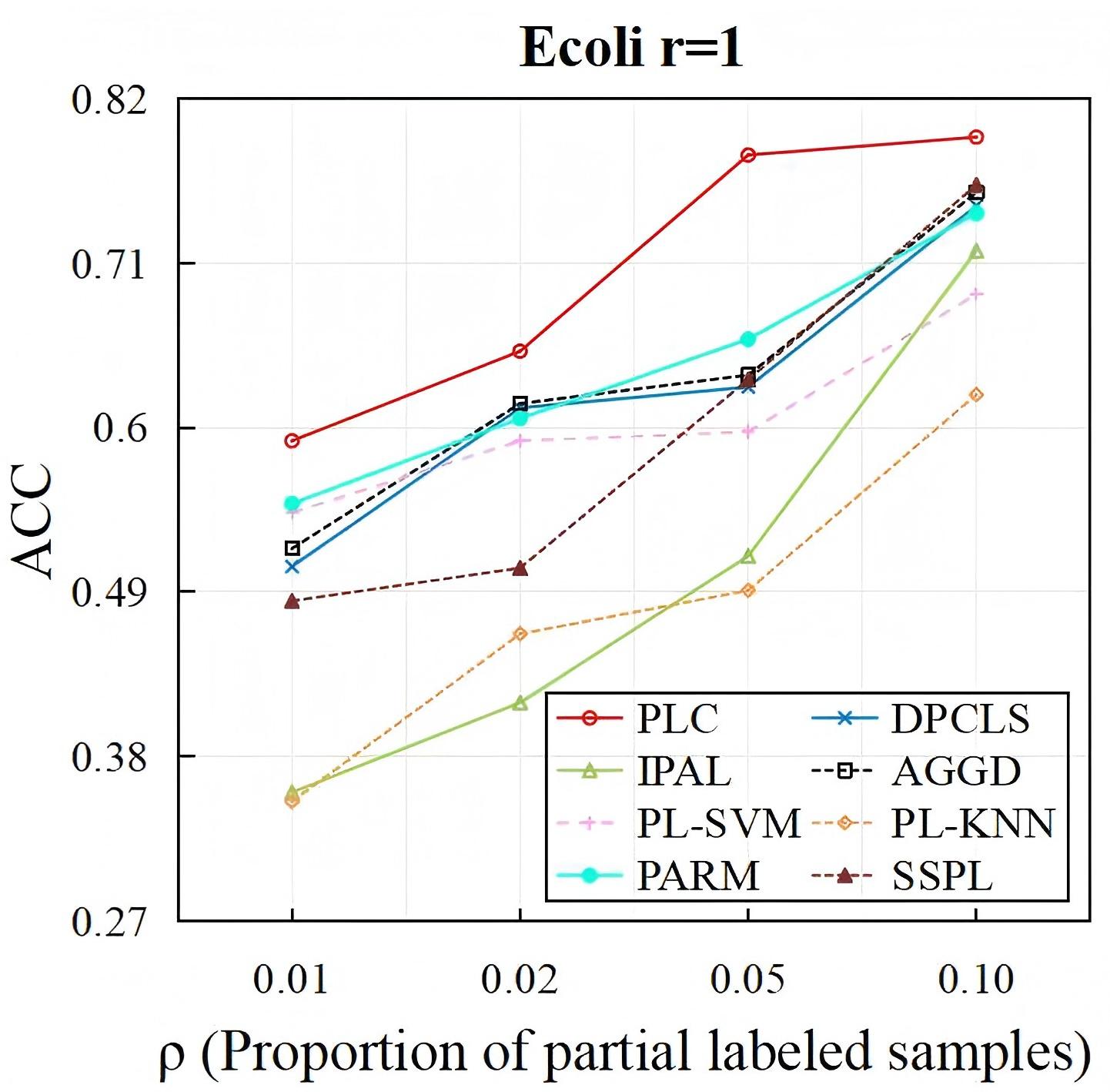} }
    \hspace{2.5mm}
    \subfigure{
    \includegraphics[width=3.5cm,height=3.5cm]{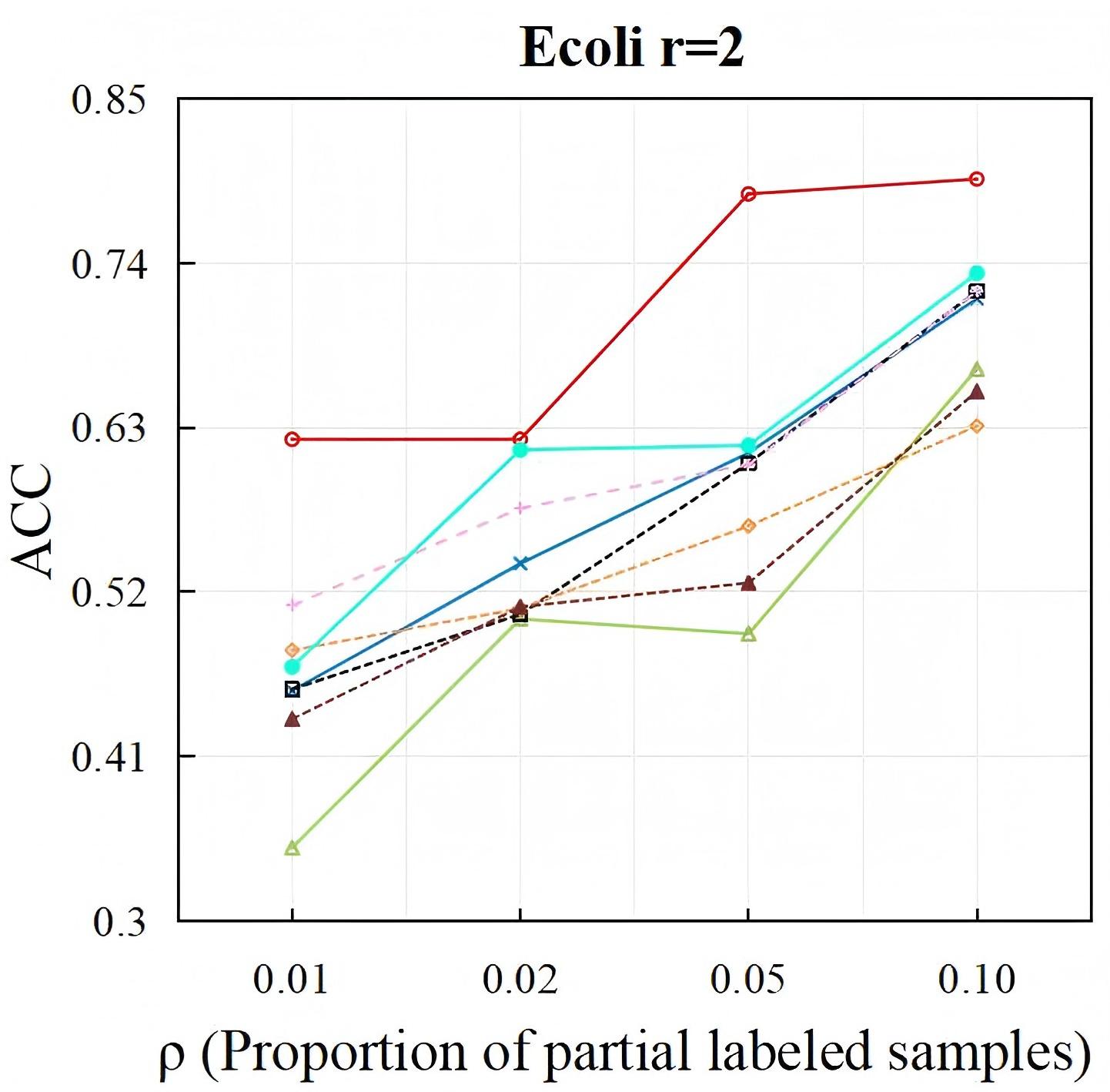} }
    \hspace{2.5mm}
    \subfigure{
    \includegraphics[width=3.5cm,height=3.5cm]{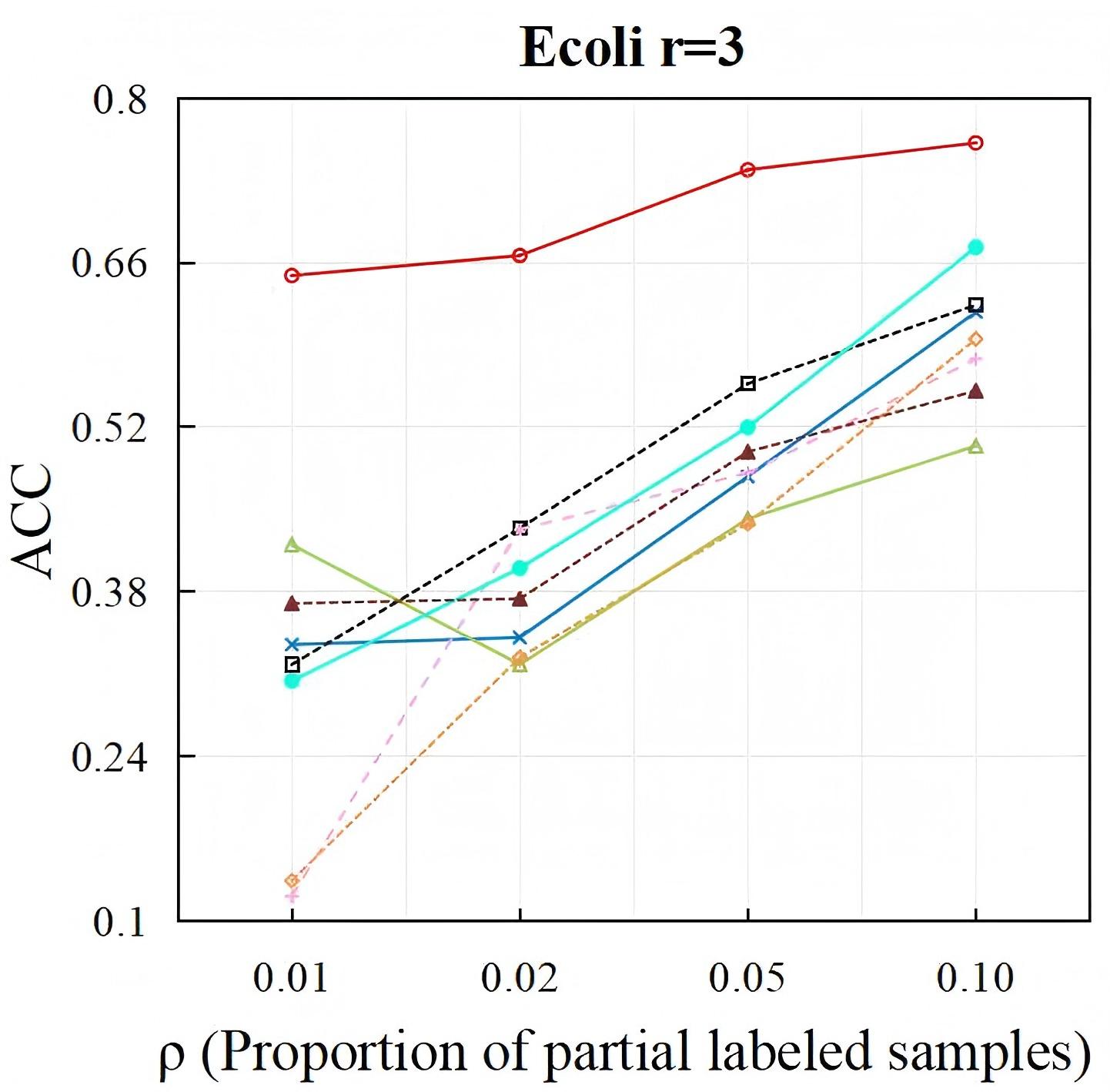} }
    \hspace{2.5mm}
    \subfigure{
    \includegraphics[width=3.5cm,height=3.5cm]{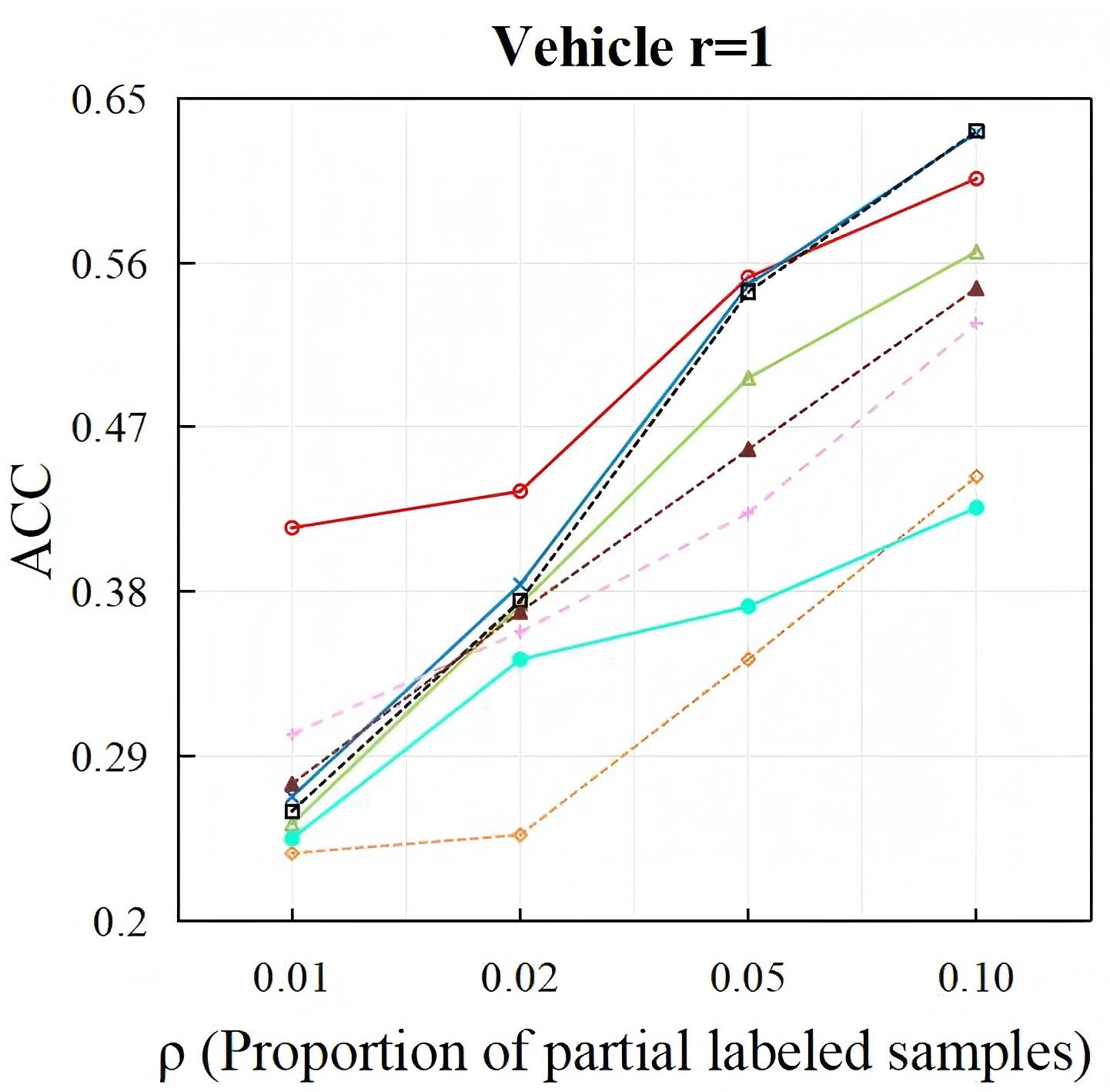} }
    \hspace{2.5mm}

    \centering
    \subfigure{
    \includegraphics[width=3.5cm,height=3.5cm]{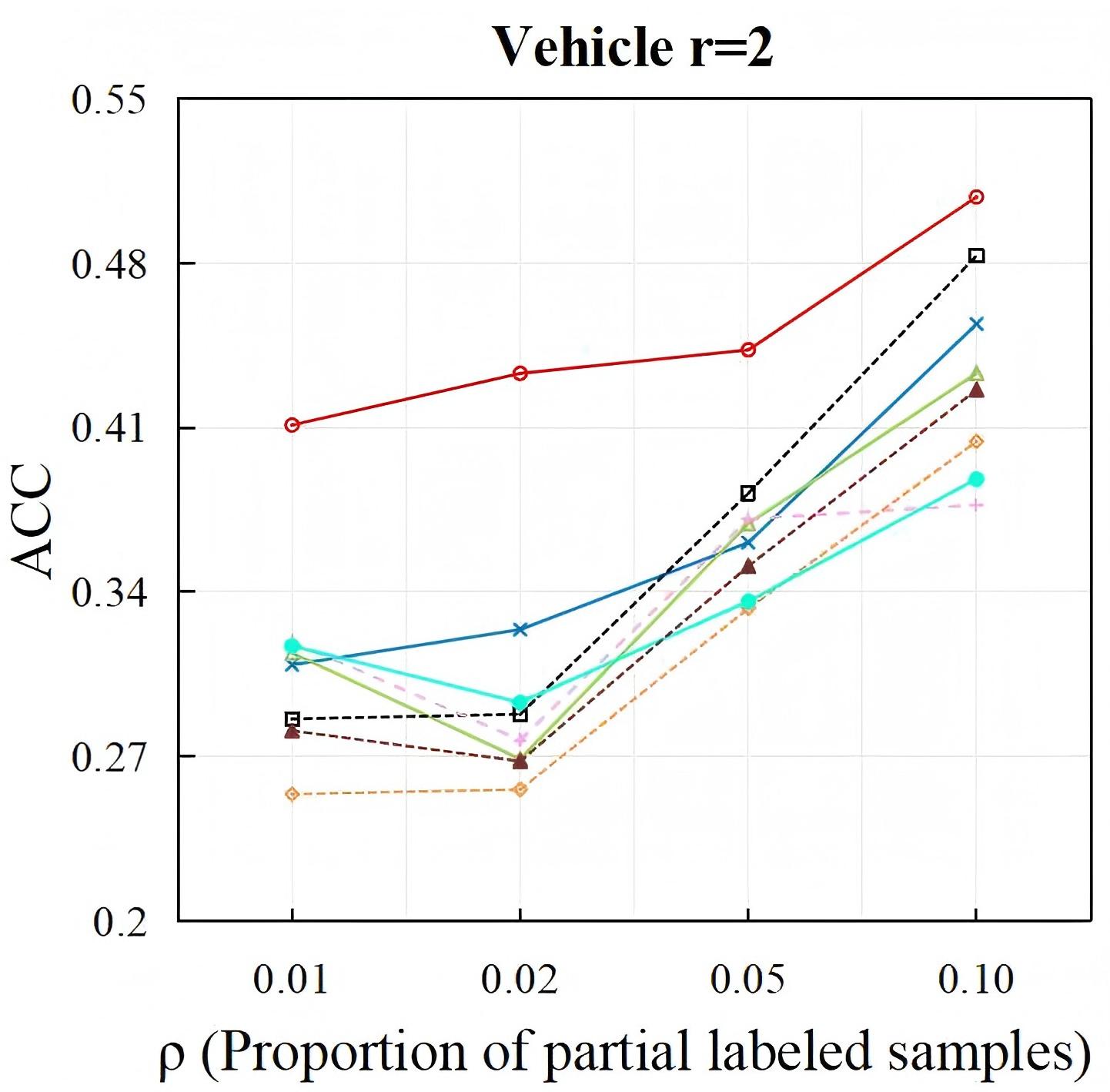} }
    \hspace{2.5mm}
    \subfigure{
    \includegraphics[width=3.5cm,height=3.5cm]{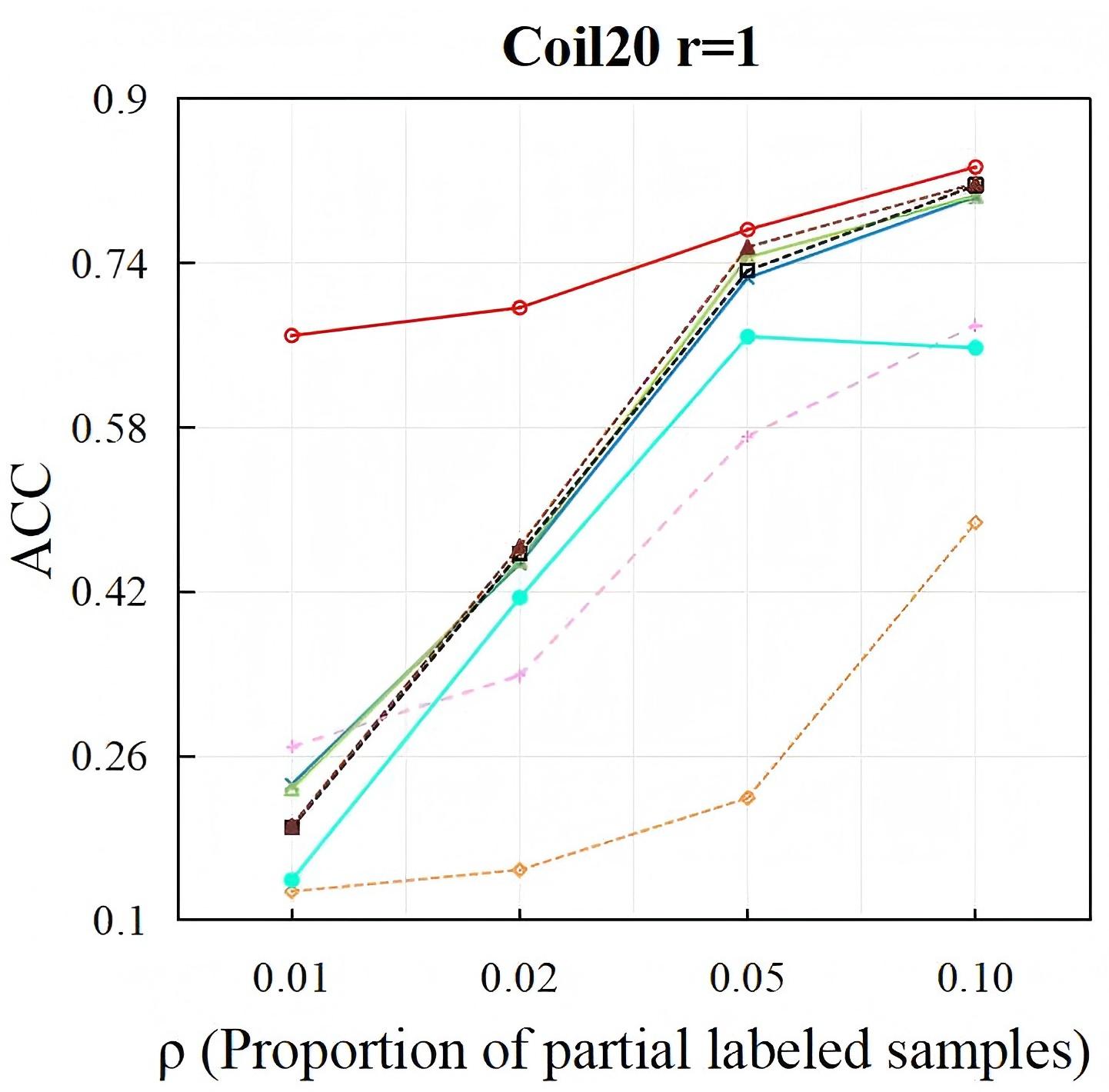} }
    \hspace{2.5mm}
    \subfigure{
    \includegraphics[width=3.5cm,height=3.5cm]{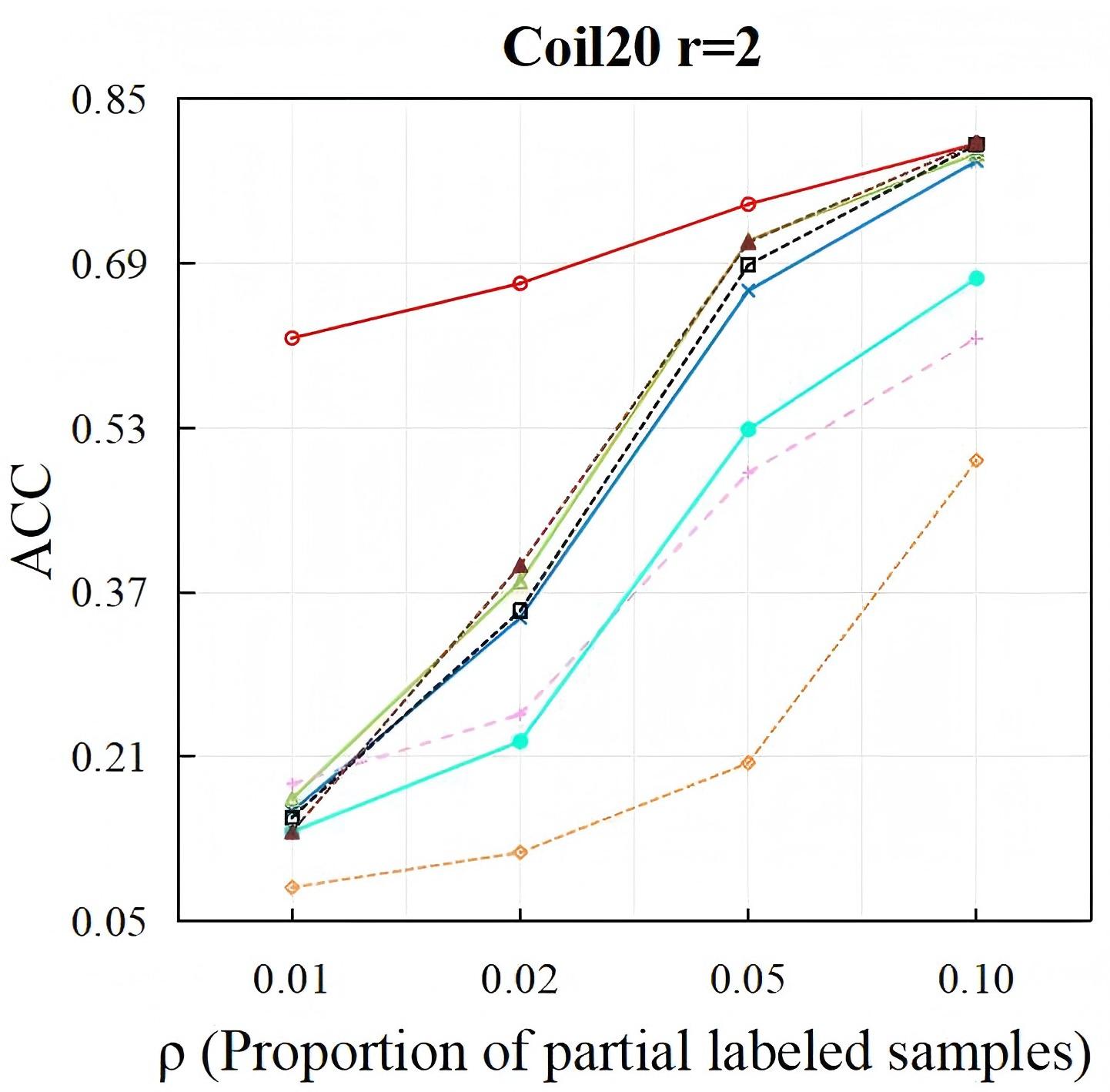} }
    \hspace{2.5mm}
    \subfigure{
    \includegraphics[width=3.5cm,height=3.5cm]{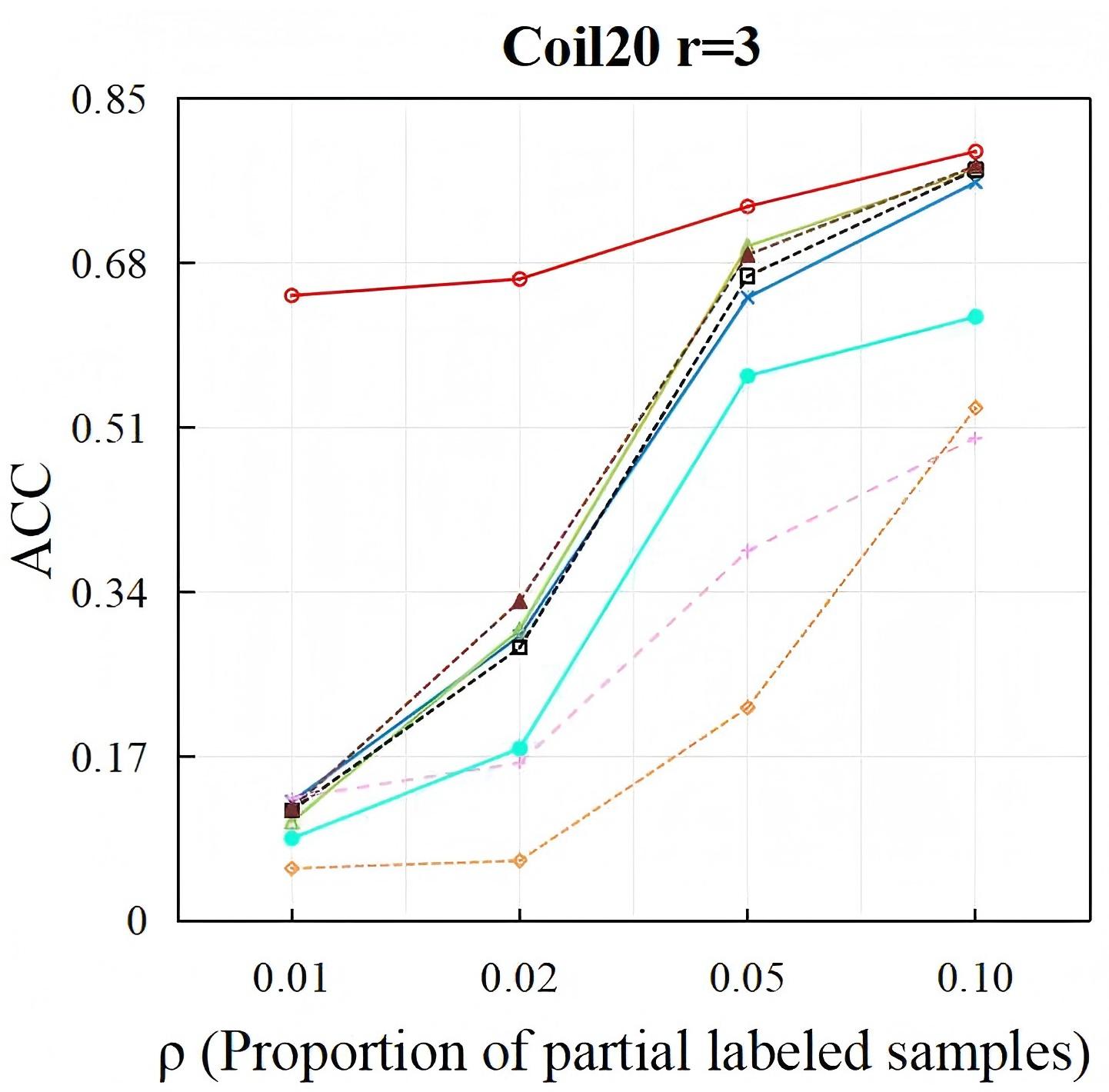} }
    \hspace{2.5mm}
    \caption{ ACCs when compared with PLL and semi-supervised PLL methods under different proportions of partial label training samples on synthetic UCI datasets. }
    \label{f4}
\end{figure*}

\begin{table*}[ht]
\centering
\begin{tabular}{ccccccc}
\Xhline{.5px}
\Xhline{.5px} 
\multicolumn{1}{l}{\multirow{2}{2cm}{\centering Compared \\ Method}} & \multicolumn{6}{c}{\textbf{LYN10}}                   \\ \cline{2-7} 
\multicolumn{1}{l}{}  & $\rho=0.05$ & $\rho=0.10$ & $\rho=0.15$ & $\rho=0.20$ & $\rho=0.30$ & $\rho=0.40$  \\ \hline
\textbf{PLC (Ours)}      & $\mathbf{0.595\pm0.007}$       & $\mathbf{0.624\pm0.008}$  & $\mathbf{0.643\pm0.011}$       & $\mathbf{0.659\pm0.007}$    & $\mathbf{0.670\pm0.010}$       & $\mathbf{0.675\pm0.009}$      \\
\textbf{K-means}     & $0.372\pm0.033$       & $0.366\pm0.022$      & $0.366\pm0.023$      & $0.371\pm0.019$  &$0.366\pm0.031$  & $0.361\pm0.018$            \\
\textbf{SC}     & $0.301\pm0.005$       & $0.299\pm0.007$      & $0.304\pm0.008$      & $0.297\pm0.006$  &$0.307\pm0.008$  & $0.302\pm0.010$            \\
\textbf{SSC-TLRR}    & $\underline{0.379\pm0.007}$       & $\underline{0.415\pm0.010}$       & $\underline{0.467\pm0.006}$       & $\underline{0.392\pm0.024}$   &$\underline{0.382\pm0.012}$ &$\underline{0.423\pm0.008}$          \\
\textbf{DP-GLPCA}   & $0.278\pm0.011$       & $0.284\pm0.009$       & $0.287\pm0.009$       & $0.287\pm0.008$ &$0.292\pm0.013$ &$0.299\pm0.012$           \\
\textbf{SSSC}   & $0.340\pm0.011$       & $0.346\pm0.009$       & $0.362\pm0.010$       & $0.358\pm0.011$ &$0.351\pm0.019$ & $0.332\pm0.011$          \\ 
\Xhline{.5px}
\Xhline{.5px} 
\end{tabular}
\caption{Experimental results on ACCs when compared with constrained clustering methods on large-scale datasets, where bold and underlined indicate the best and second best results respectively.}
\label{tb7}
\end{table*}

\begin{table*}[ht]
\centering
\begin{tabular}{ccccccccccc}
\Xhline{.5px}
\hline
\textbf{}                 & \textbf{DPCLS}              & \textbf{AGGD}               & \textbf{IPAL}              & \textbf{PL-SVM} & \textbf{PL-KNN} & \textbf{SSPL} & \textbf{PARM} & \textbf{SSC-TLRR} & \textbf{DP-GLPCA} & \textbf{SSSC} \\ \hline
(\romannumeral1)                       & 11/6/1                      & 11/7/1                      & \multicolumn{1}{c}{14/4/0} & 17/1/0          & 18/0/0          & 16/2/0        & 17/1/0        & 30/0/0            & 30/0/0            & 30/0/0        \\
(\romannumeral2)                       & 28/4/0                      & 25/6/1                      & \multicolumn{1}{c}{29/3/0} & 32/0/0          & 32/0/0          & 27/5/0        & 32/0/0        & 31/17/0           & 48/0/0            & 35/13/0       \\
\multicolumn{1}{l}{Total} & \multicolumn{1}{l}{39/10/1} & \multicolumn{1}{l}{36/13/1} & 43/7/0                     & 49/1/0          & 50/0/0          & 43/7/0        & 49/1/0        & 61/17/0           & 78/0/0            & 65/13/0       \\ \hline
\Xhline{.5px}
\end{tabular}
\caption{Win/tie/loss counts on the classification performance of PLC against the PLL, semi-supervised PLL and constrained clustering methods on all datasets. (\romannumeral1), (\romannumeral2) indicate the summaries on real-world datasets and synthetic UCI datasets. "Total" denotes the summary on all the datasets.}
\label{tb12}
\end{table*}

\begin{table}[ht]
\centering
\begin{tabular}{ccc}
\Xhline{.5px}
\Xhline{.5px} 
\multicolumn{1}{l}{\multirow{2}{2cm}{\centering Compared \\ Method}} & \multicolumn{2}{c}{\textbf{LYN10}}                   \\ \cline{2-3} 
\multicolumn{1}{l}{}  & $\rho=0.01$ & $\rho=0.02$   \\ \hline
\textbf{PLC (Ours)}  & $\mathbf{0.525\pm0.024}$     & $\mathbf{0.556\pm0.010}$       \\
\textbf{DPCLS}     & $0.485\pm0.016$       & $0.523\pm0.020$        \\
\textbf{AGGD}     & $0.493\pm0.016$       & $0.538\pm0.019$  \\
\textbf{IPAL}    & $0.468\pm0.015$       & $0.522\pm0.013$        \\
\textbf{PL-kNN}   & $0.394\pm0.025$       & $0.426\pm0.016$    \\
\textbf{PL-SVM}   & $0.485\pm0.038$       & $0.542\pm0.016$    \\
\textbf{PARM}   & $\underline{0.497\pm0.010}$       & $\underline{0.550\pm0.018}$        \\ 
\textbf{SSPL}   & $0.445\pm0.041$       & $0.482\pm0.021$    \\
\Xhline{.5px}
\Xhline{.5px} 
\end{tabular}
\caption{ACCs when compared with PLL and semi-supervised PLL methods on large-scale dataset, where bold and underlined indicate the best and second best results respectively.}
\label{tb3}
\end{table}

\section{Complexity Analysis} The computational complexity of our algorithm is dominated by steps 7-11. In steps 7-9, we use interior point method \cite{Ye1989AnEO} to solve a series of QP problems with the complexity of $\mathcal{O}(nk^3)$. Similarly, step 10 solves a QP problem with the complexity of $\mathcal{O}(n^3q^3)$. When dealing with large datasets, we can transform the original problem into a series of subproblems as Eq. (9) with the complexity of $\mathcal{O}(nq^3)$. Step 11 solves the problem by KKT conditions with the complexity of $\mathcal{O}(n^3)$. In summary, the overall complexity of our algorithm in each iteration is $\mathcal{O}(nk^3+n^3q^3+n^3)$ and $\mathcal{O}(nk^3+nq^3+n^3)$ for large datasets.


\section{Details of Compared Datasets}
Table \ref{tb1} summarizes the characteristics of controlled UCI datasets and real-world datasets. Following the widely-used partial label data generation protocol \cite{2011Learning}, we generate artificial partial label datasets under the parameter $r$ which controls the number of false-positive labels\footnote{For Vehicle, the setting $r=3$ is not considered as there are only four class labels in the label space}. 

The real-world datasets are collected from various domains including Lost \cite{2009Learning} for automatic face naming, MSRCv2 \cite{2014A} for object classification, Mirflickr \cite{2008The} for web image classification and BirdSong \cite{2012Rank} for bird song classification.

\section{Supplementary Experimental Results}
\subsection{Comparison to Constrained Clustering}
Fig. \ref{f2} illustrates the ACCs and NMIs of our PLC method compared to the constrained clustering methods on the datasets Ecoli $r=3$ and Coil20 $r=3$. Our PLC method ranks first in 87.5\% (21/24) cases which further proves the effectiveness of our PLC method.

\subsection{Comparison to PLL \& Semi-supervised PLL}
Fig. \ref{f4} illustrates the ACCs of our PLC method compared to the PLL and semi-supervised PLL methods on synthetic UCI datasets. Our PLC method achieves superior or competitive performance against the comparing methods with a lower proportion of partial labeled samples. According to Fig. \ref{f4}, PLC method achieves superior performance against PL-KNN and PL-SVM in 100\% (32/32) cases, against IPAL, DPCLS, PARM and SSPL in 96.88\% (31/32) cases, and against AGGD in 93.75\% (30/32) cases. The experimental results further prove that our PLC method performs well in the case of fewer partial label training samples.

\subsection{Experiment on Large-scale Dataset}
LYN10 is a large-scale dataset for automatic face naming, consisting of samples from the top 10 classes of the Yahoo!News \cite{Guillaumin2010MultipleIM} dataset. The characteristics of LYN10 are shown in Table \ref{tb1}. Table \ref{tb7} reports the ACCs of our PLC method compared to constrained clustering methods on the large-scale dataset. Table \ref{tb3} reports the ACCs of our PLC method compared to PLL and semi-supervised PLL methods on the large-scale dataset. Our PLC method performs well on the large-scale dataset and ranks first in all experimental settings.

\subsection{Significance Analysis}
Table \ref{tb12} reports win/tie/loss counts between our PLC methods and ten comparing methods on the real-world datasets and synthetic UCI datasets according to the pairwise t-test at the significance level of 0.05. We can find that our PLC method statistically outperforms the PLL and semi-supervised PLL methods (the first seven columns) in 88.3\% (309/350) cases and statistically outperforms the constrained clustering methods (the last three columns) in 87.2\% (204/234) cases.


\end{document}